%% file: main.tex
\documentclass[10pt,twocolumn,letterpaper]{article}

\usepackage[pagenumbers]{cvpr} 


\definecolor{cvprblue}{rgb}{0.21,0.49,0.74}
\usepackage[bookmarks=false,pagebackref,breaklinks,colorlinks,allcolors=cvprblue]{hyperref}
\usepackage{pifont}
\usepackage{multirow}
\usepackage[dvipsnames]{xcolor}
\usepackage[accsupp]{axessibility}
\usepackage{bbm}
\usepackage{placeins}

\newcommand{\return}{G}

\newcommand{\jointstate}{s}

\newcommand{\returnset}{\mathbb{G}}
\newcommand{\jointactionset}{\mathbb{A}}
\newcommand{\jointstateset}{\mathbb{S}}

\newcommand{\std}[1]{\textcolor{black}{\scriptsize{$\pm #1$}}}

\definecolor{custom_red}{RGB}{222, 89, 89}
\definecolor{custom_blue}{RGB}{135, 179, 230}
\definecolor{custom_purple}{RGB}{190, 169, 245}

\def\paperID{14074} 
\def\confName{CVPR}
\def\confYear{2025}

\title{Scenario Dreamer:\\ Vectorized Latent Diffusion for Generating Driving Simulation Environments}

\author{Luke Rowe $^{1,2,6}$, Roger Girgis$^{1,3,6}$, Anthony Gosselin$^{1,3}$, \\ Liam Paull$^{1,2,5}$, Christopher Pal$^{1,2,3,5}$, Felix Heide$^{4,6}$\\
$^1$Mila, $^2$Université de Montréal, $^3$Polytechnique Montréal, \\$^4$Princeton University, 
$^5$CIFAR AI Chair, $^6$Torc Robotics \\ {\normalsize \textbf{\url{https://princeton-computational-imaging.github.io/scenario-dreamer}}}
}

\makeatletter
\newcommand{\@LN}[2]{} 
\makeatother

\begin{document}
\maketitle
\input{sec/0_abstract}
\input{sec/1_introduction} 
\input{sec/2_related_work}
\input{sec/3_background}

\input{sec/4_method}
\input{sec/5_experiments}
\input{sec/6_conclusion}

\clearpage
\section{Acknowledgements}

Felix Heide was supported by an NSF CAREER Award (2047359), a Packard Foundation Fellowship, a Sloan Research Fellowship, a Sony Young Faculty Award, a Project X Innovation Award, and an Amazon Science Research Award. C. Pal thanks the IVADO, the NSERC Discovery Grants and the CIFAR AI Chairs programs for their support.

{
    \small
    \bibliographystyle{ieeenat_fullname}
    \bibliography{main}
}

\onecolumn
\appendix
\clearpage

\section{Additional Results}
\input{supplemental/results}
\section{Simulation Framework Details}
\input{supplemental/simulation}
\section{Model Details}
\input{supplemental/model}
\section{Evaluation Details}
\input{supplemental/evaluation}

\end{document}

%% file: sec/0_abstract.tex
\begin{abstract}

We introduce Scenario Dreamer, a fully data-driven generative simulator for autonomous vehicle planning that generates both the initial traffic scene—comprising a lane graph and agent bounding boxes—and closed-loop agent behaviours. Existing methods for generating driving simulation environments encode the initial traffic scene as a rasterized image and, as such, require parameter-heavy networks that perform unnecessary computation due to many empty pixels in the rasterized scene. Moreover, we find that existing methods that employ rule-based agent behaviours lack diversity and realism. Scenario Dreamer instead employs a novel vectorized latent diffusion model for initial scene generation that directly operates on the vectorized scene elements and an autoregressive Transformer for data-driven agent behaviour simulation. Scenario Dreamer additionally supports scene extrapolation via diffusion inpainting, enabling the generation of unbounded simulation environments. Extensive experiments show that Scenario Dreamer outperforms existing generative simulators in realism and efficiency: the vectorized scene-generation base model achieves superior generation quality with around 2× fewer parameters, 6× lower generation latency, and 10× fewer GPU training hours compared to the strongest baseline. We confirm its practical utility by showing that reinforcement learning planning agents are more challenged in Scenario Dreamer environments than traditional non-generative simulation environments, especially on long and adversarial driving environments.
\end{abstract}



%% file: sec/1_introduction.tex
\section{Introduction}

Simulators are invaluable tools for the safe and scalable development of autonomous vehicles (AVs) \citep{scanlon2021waymo, yang2023unisim, chen2024trafficsimsurvey, li2024choosesim}. They reduce costs and accelerate development, while offering a safe way to evaluate the performance of autonomous vehicle systems under diverse, rare, and potentially dangerous real-world conditions. Despite significant advancements in enhancing the speed and realism of autononous driving simulators \citep{caesar2024nuplan, eugene2024gpudrive, gulino2023waymax}, current simulators remain constrained by their reliance on prerecorded driving logs. Specifically, current data-driven simulators either replay or modify existing driving logs, limiting scalability due to the finite size and diversity of the available data. For instance, public driving simulators like GPUDrive \citep{eugene2024gpudrive} and Waymax \citep{gulino2023waymax}, which are based on the Waymo Open Motion Dataset \citep{ettinger2021womd}, cover only 1,750 km of unique roadway—a fraction of the nearly 22,000 km of roadway that an average human drives annually \citep{averagemiles}. Simulations are also bounded by the length of the driving logs, which are typically less than 30 seconds. Given the billions of miles of diverse driving experience needed to validate the safety of autonomous vehicles in simulation, existing simulation environments derived from traffic logs have struggled to enable scalable training and evaluation of autonomous vehicles.

To address the limitations of current driving simulators, recent work has shown that diffusion generative models can be used to synthesize driving environments in an abstract representation comprising bird's eye view (BEV) agent bounding boxes and a lane graph \citep{chitta2024sledge, sun2024drivescenegen}. However, these methods are either not fully data-driven as the agent behaviours are governed by unrealistic rule-based models \citep{chitta2024sledge}, or can only generate short driving scenarios with non-reactive agents \citep{sun2024drivescenegen}. Prior methods also encode driving scenes as rasterized BEV images, which require costly and unnecessary computation due to many empty pixels in the rasterized scenes \citep{sun2024drivescenegen, chitta2024sledge}. While existing generative simulators from sensor data have been proposed \citep{yang2024drivearena}, these methods fail to meet the strict latency demands of simulators due to their reliance on large video diffusion models.

We propose Scenario Dreamer, a fully data-driven \textit{generative} simulator for autonomous driving planning. Scenario Dreamer decomposes environment generation into \textit{initial scene generation}, which generates the initial scene in an abstract representation comprising BEV agent bounding boxes and a lane graph, and \textit{behaviour simulation}, which controls agent behaviours over time. A core component of Scenario Dreamer is our proposed vectorized latent diffusion model for initial scene generation that \emph{operates directly on the vectorized scene elements.} Compared to a rasterized encoding, the proposed model offers several practical advantages, including higher fidelity generations 
with reduced generation latency and higher quality scene extrapolations even at dense and complicated road geometries. It additionally enables learning the lane graph connectivity explicitly by modeling pairwise relationships between road elements, thus eliminating the need for post-processing heuristics used by \citet{chitta2024sledge}. For closed-loop behaviour simulation, we utilize a return-conditioned autoregressive Transformer \citep{rowe2024ctrlsim} for controllable agent behaviours that is adapted to support longer simulation rollouts. Both models are trained from real-world driving data. 


Unlike prior work, Scenario Dreamer supports flexible test-time control over the generation of challenging and safety-critical scenarios. Specifically, we can explicitly control scene density with Scenario Dreamer by specifying the number of lanes and agents within a given field of view (FOV). Moreover, we can simulate adversarial agent behaviours with Scenario Dreamer via exponential tilting of the behaviour model \citep{rowe2024ctrlsim, lee2022multigamedt}. We integrate Scenario Dreamer with GPUDrive \citep{eugene2024gpudrive} to enable training Scenario Dreamer compatible reinforcement learning (RL) agents and evaluate RL agents in Scenario Dreamer environments. We show that Scenario Dreamer provides more challenging environments for RL planning agents than existing non-generative environments. Scenario Dreamer unlocks the ability to synthesize a limitless quantity of unbounded and interactive safety-critical environments, which we hope stimulates future autonomous driving planning research. 

\textbf{Contributions}: (1) We introduce Scenario Dreamer, a fully data-driven \textit{generative} simulator for autonomous driving planning. At the core of Scenario Dreamer is a novel vectorized latent diffusion model for initial scene generation that offers practical advantages over prior methods that utilize a rasterized scene encoding. (2) We show that Scenario Dreamer environments challenge RL planners, especially on long and adversarial driving environments.

%% file: sec/2_related_work.tex
\section{Related Work}

\textbf{Simulators for Autonomous Driving} Traditional handcrafted driving simulators require costly manual effort to create environments \citep{codevilla2017carla, highway-env, cai2020summit}. Consequently, they are difficult to scale to larger environments and suffer from a significant sim-to-real gap. To address these issues, recent works have proposed data-driven driving simulators that are based on real driving data \citep{chen2024trafficsimsurvey}. These simulators either reconstruct driving environments from sensor inputs using neural rendering techniques \citep{yang2023unisim, yang2023emernerf, yan2024street, ljungbergh2024neuroncap, zhou2024drivinggaussian, zhou2024hugs, zhou2024hugsim, xie2023snerf, chen2025snerf, tonderski2024neurad, ost2021nsg}, or they utilize post-perception outputs to represent driving scenes in an abstract representation comprising agent bounding boxes and HD maps \citep{vinitsky2022nocturne, li2023metadrive, gulino2023waymax, lavington2024torchdrivenv, caesar2024nuplan, eugene2024gpudrive, zhou2020smarts}.  Scenario Dreamer adopts the abstract environment representation due to its lower computational demands and memory footprint, while offering sufficient complexity to evaluate AV planners.

\textbf{Generative Simulation Environments} While open-source data-driven simulators reduce the sim-to-real gap compared to handcrafted simulators, they are fundamentally limited by the scarcity of public driving data and the short durations of prerecorded driving logs. To overcome these constraints, recent works have proposed using diffusion generative models to generate driving simulation environments from sensor inputs \citep{hu2023gaia-1, yang2024drivearena, gao2024vista} or post-perception data \citep{sun2024drivescenegen, chitta2024sledge}. A similar line of work uses diffusion models as generative interactive environments in other contexts, such as for video games \citep{bruce2024genie, alonso2024worldmodel, valevski2024gamengen} and robotics \citep{yang2024unisim, yu2024adaptive, wu2024ivideogpt}. SLEDGE employs a customized latent diffusion model to generate bounding boxes and lane graphs but relies on rule-based agent behaviours \citep{treiber2000idm, kesting2007mobil}, limiting realism \citep{chitta2024sledge}. DriveSceneGen uses a diffusion model for initial scene generation and a motion prediction model for agent behaviours but produces only short, non-reactive scenarios of up to 20 seconds \citep{sun2024drivescenegen}. Scenario Dreamer addresses these shortcomings by introducing a latent diffusion model for initial scene generation and a Transformer-based driving policy \citep{rowe2024ctrlsim} for closed-loop behaviour simulation, making it a fully data-driven, closed-loop, generative driving simulator capable of generating arbitrarily long simulations. 

\textbf{Initial Scene Generation and Behaviour Simulation} A large body of work focuses specifically on the task of initial scene generation, targeting either the generation of initial agent bounding boxes \citep{tan2021scenegen, lu2024scenecontrol}, lane graphs \citep{mi2021hdmapgen}, or both \citep{chitta2024sledge, sun2024drivescenegen}. SceneControl \citep{lu2024scenecontrol} utilizes a vectorized diffusion model for initial agent generation but does so in data space. Both SLEDGE and DriveSceneGen employ diffusion models to generate the initial scene by processing it as a rasterized image, leading to unnecessary computation. In contrast, Scenario Dreamer generates the initial scene with vectorized processing \citep{gao2020vectornet, hoogeboom2022edm, xu2023geoldm, chen2023polydiffuse, jiang2023vad, chen2024vadv2, ngiam2022scenetransformer, girgis2021autobots, rowe2023fjmp, zhou2022hivt, liao2023maptr, liao2025maptrv2, jiang2023motiondiffuser}, which enhances the speed and performance of the scene generations and extrapolations.

\begin{figure*}[tb]
\vspace{-20pt}
    \centering
    \begin{subfigure}[b]{0.5\textwidth}
        \centering
        \includegraphics[width=\textwidth]{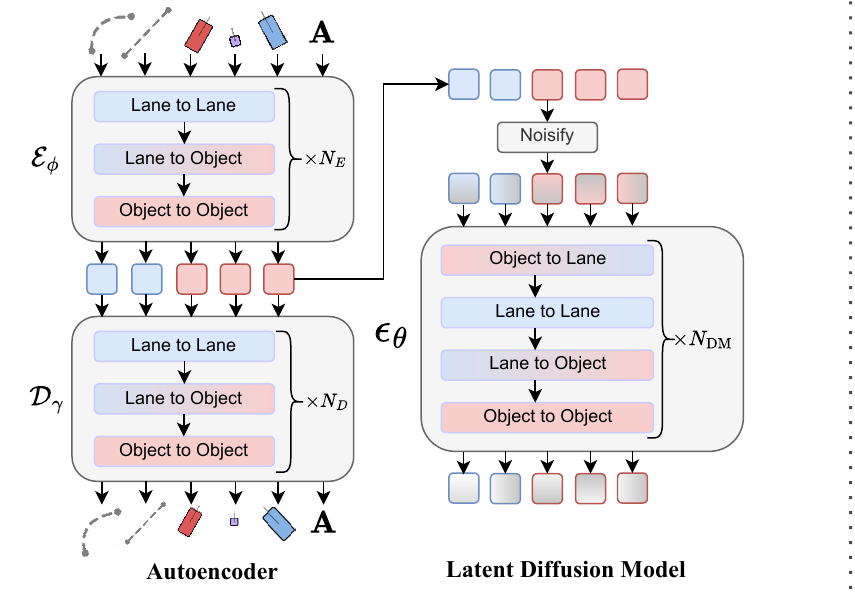}
        \caption{\textbf{Model Overview}}
        \label{fig:method-overview}
    \end{subfigure}
    \begin{subfigure}[b]{0.4\textwidth}
        \centering
        \raisebox{1.5cm}{\includegraphics[width=\textwidth]{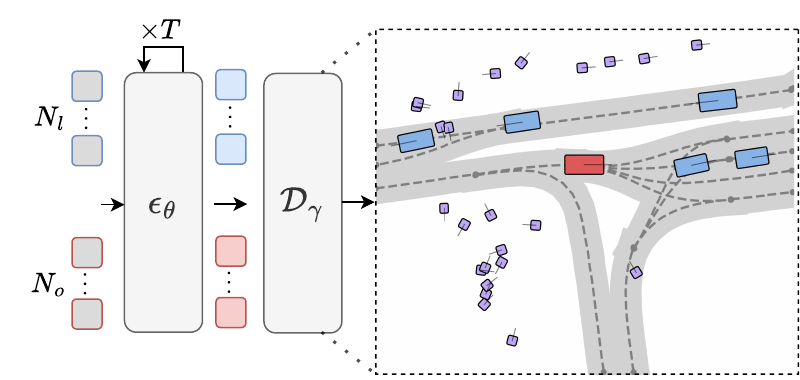}}
        \caption{\textbf{Inference Time Sampling}}
        \label{fig:method-inf} 
    \end{subfigure} \vspace{-5pt}
    \caption{\textbf{Scenario Dreamer vectorized latent diffusion model for initial scene generation.} \textit{Left}: We embed each vectorized scene element into a latent representation with an autoencoder parameterized with factorized attention blocks, which additionally fuses the lane connectivity $\mathbf{A}$. The latent Transformer diffusion model $\epsilon_{\theta}$ is trained to sample from the autoencoder's latent distribution. \textit{Right}: Scenario Dreamer samples novel driving scenes by initializing $N_o + N_l$ noise tokens which are iteratively denoised with $\epsilon_{\theta}$ over $T$ steps and decoded into vectorized scene elements. The ego vehicle is denoted in \color{custom_red}red\color{black}, with other agents colored in \color{custom_blue}blue \color{black} and pedestrians in \color{custom_purple}purple\color{black}.}
    \label{fig:method}\vspace{-15pt}
\end{figure*}

Recent works have proposed various methods for simulating closed-loop agent behaviours conditioned on initial agent states and HD maps \citep{suo2021trafficsim, igl2022symphony, xu2023bits, suo2023mixsim, zhang2023trafficbots, cornelisse2024hrppo, zhou2024behaviorgpt}. These approaches rely on large Transformer-based models \citep{philion2024trajeglish, hu2024gump, wu2024smart} or diffusion models \citep{zhong2023ctg, zhong2023ctgpluspluslanguage, huang2024versatile, guo2023scenedm, jiang2024scenediffuser}, which are computationally expensive, especially considering that behaviour models run autoregressively in the loop during simulation. Scenario Dreamer utilizes a lightweight return-conditioned behaviour model \citep{rowe2024ctrlsim} for closed-loop behaviour simulation. The return conditioning allows for flexible test-time control over agent behaviours, which can be used to generate adversarial driving behaviours. Researchers have also explored methods that attempt to generate both initial agent states and behaviours conditioned on the map and ego state, either by using separate modules for each task \citep{bergamini2021simnet, feng2023trafficgen} or by coupling them in an end-to-end trainable system \citep{tan2023language, whiteson2024unigen, pronovost2023scenariodiffusion, jiang2024scenediffuser}. Scenario Dreamer keeps initial scene generation and behaviour simulation decoupled by design, allowing for these components to be invoked at different frequencies as needed.

%% file: sec/3_background.tex

%% file: sec/4_method.tex
\section{Scenario Dreamer}

We approach the task of generative driving simulation by decomposing it into \textit{initial scene generation} and \textit{behaviour simulation}. Below, we detail our proposed initial scene generator and agent behaviour model, both trained from real-world driving data. We then describe our generative simulation framework that supports closed-loop evaluation of AV planning agents. 

\subsection{Problem Setting}  

\begin{figure}
  \centering
\includegraphics[width=0.9\columnwidth]{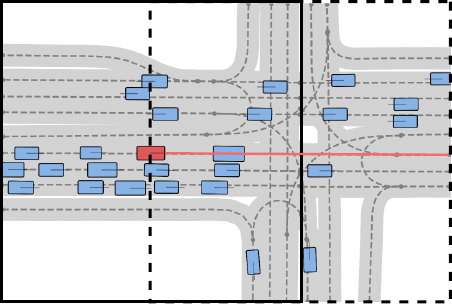}
    \caption{Illustrative example of Scenario Dreamer's inpainting capabilities, where the initial tile is outlined in solid lines and the inpainted tile in dashed. The model generates consistent lane geometries at the scene boundaries, even at complex intersections.}
    \label{fig:inpainting}
    \vspace{-10pt}
\end{figure}

\textbf{Initial Scene Generation} Initial scene generation involves generating the initial BEV object bounding box states and the underlying map structure within a fixed field of view (FOV). Following \citep{chitta2024sledge}, we generate a $64$m$\times64$m FOV centered and rotated to the ego agent. We denote by $F$ the $64$m$\times64$m region being generated and denote by $F_P$ and $F_N$ the $32\text{m} \times 64\text{m}$ regions that are ahead of and behind the ego agent, respectively. The initial scene generator is tasked with sampling from the distribution $p(\mathcal{I}_F)$ of initial scenes, where an initial scene $\mathcal{I}_F = \{ \mathcal{O}, \mathcal{M} \}$ comprises a set of objects $\mathcal{O}$ and the map structure $\mathcal{M}$ within the FOV $F$. We define $\mathcal{O} = \{\mathbf{o}_i\}_{i=1}^{N_o}$ as a set of $N_o$ objects that includes the traffic participants (\textit{e.g.,} ego agent, vehicles, pedestrians, cyclists) and static objects (\textit{e.g.,} traffic cones), where $\mathbf{o}_i$ is an 8-dimensional vector containing the 2-dimensional position, speed, cosine and sine of the heading, length, width, and object class of bounding box $i$. We generate a map representation $\mathcal{M}$ similar to \citep{chitta2024sledge}, where $\mathcal{M} = \{ \mathcal{L}, \mathbf{A} \}$ contains a set $\mathcal{L} = \{ \mathbf{l}_i \}_{i=1}^{N_l}$ of $N_l$ centerlines where each $\mathbf{l}_i$ is a $20\times2$ sequence of centerline positions, and $\mathbf{A} \in \{0,1\}^{N_l \times N_l \times 4}$ defines the associated centerline connectivity as a stack of four adjacency matrices describing the successor, predecessor, left, and right neighbor connections. The initial scene generator must also support sampling from the conditional distribution $p(\mathcal{I}_{F_P} | \mathcal{I}_{F_N})$ as this enables sampling arbitrarily long scenes by stitching together new generated regions $F_P$ conditioned on existing regions $F_N$ (see Figure \ref{fig:inpainting}).

\textbf{Behaviour Simulation} Given an initial scene configuration prescribed by the initial scene generator, the task of behaviour simulation involves modeling the behaviour of the dynamic objects in the scene over time. Concretely, given the set of initial object bounding box states $\jointstateset_{0} := \mathcal{O}$ and the lane structure $\mathcal{M}$, the behaviour model employs a multi-agent driving policy $\pi(\jointactionset_t | \jointstateset_t, \mathcal{M})$ and a forward transition model $\mathcal{P}(\jointstateset_{t+1} | \jointstateset_{t}, \jointactionset_{t})$, where $\jointactionset_{t}$ is set of the actions of all dynamic objects at timestep $t$.

\begin{figure*}
\vspace{-20pt}
  \centering
  \includegraphics[width=\textwidth]{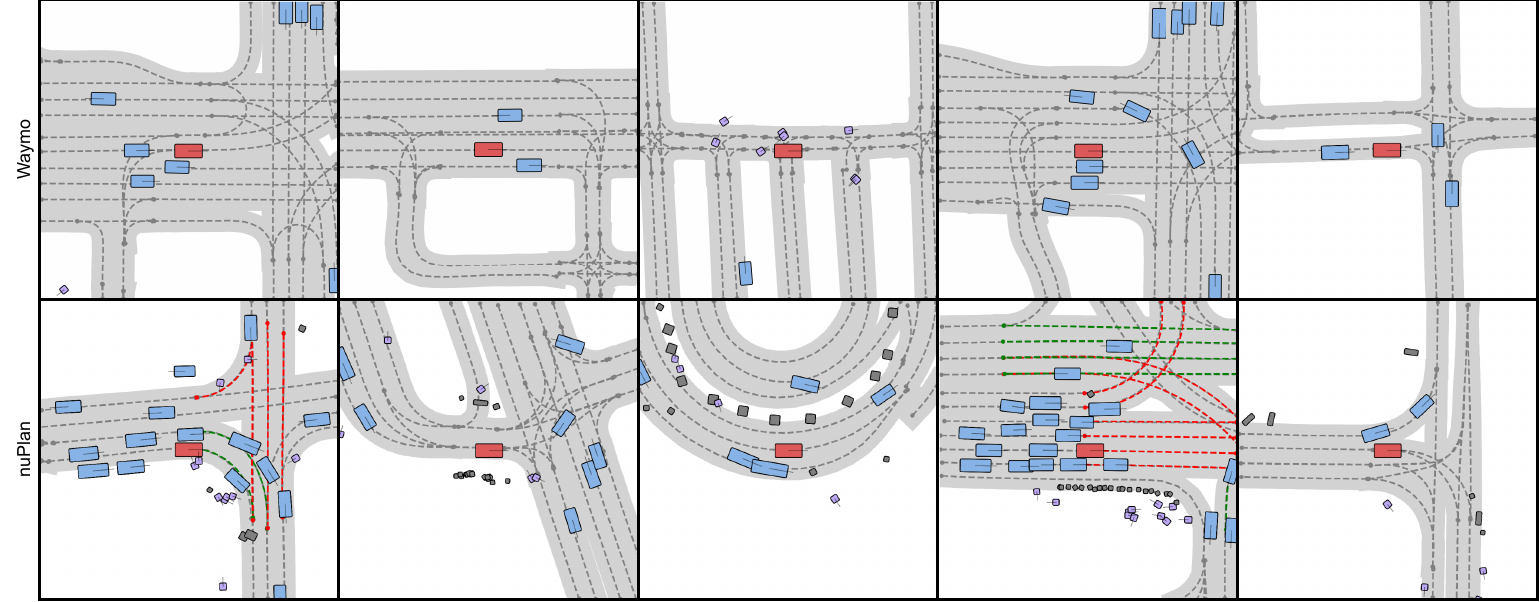}
    \caption{Vectorized environments generated by Scenario Dreamer with the proposed vectorized latent diffusion model trained on the Waymo dataset (top row) and nuPlan dataset (bottom row).}
    \label{fig:qualitative}
    \vspace{-10pt}
\end{figure*}


\subsection{Vectorized Latent Diffusion Model}
\label{subsec:latent-diffusion-model}

The initial scene generator aims to sample from $p(\mathcal{I}_F)$. To achieve this, we propose learning an approximation, $\tilde{p}(\mathcal{I}_F) \approx p(\mathcal{I}_F)$, based on real driving data with a diffusion model, given the strong capability of diffusion models to capture highly complex distributions \citep{ramesh2022dalle, Abramson2024}. We employ a two-stage training process: first, we learn a compact latent representation of the $N_o$ objects, $\{\mathbf{h}^{\mathcal{O}}_i\}_{i=1}^{N_o}$, and the $N_l$ centerlines, $\{\mathbf{h}^{\mathcal{L}}_i\}_{i=1}^{N_l}$, through a low-$\beta$ variational autoencoder \citep{higgins2017betavae}; next, we train a diffusion model to sample from the latent distribution $p(H) := p(\{\mathbf{h}^{\mathcal{O}}_i\}_{i=1}^{N_o}, \{\mathbf{h}^{\mathcal{L}}_i\}_{i=1}^{N_l})$. The architecture is illustrated in Figure \ref{fig:method}, and detailed below.

\vspace{-12pt}
\subsubsection{Autoencoder}
\vspace{-6pt}
The autoencoder consists of a Transformer-based \citep{vaswani2017attention} encoder $\mathcal{E}_{\phi}$ and decoder $\mathcal{D}_{\gamma}$ that operate directly on the vectorized scene elements -- in contrast, prior work \citep{chitta2024sledge} encodes the driving scene as a rasterized image.

\textbf{Encoder} The encoder first embeds the $N_o + N_l$ vector elements with a per-vector MLP and additionally embeds the one-hot representation of the lane connectivity type $c_{ij}$ with an MLP for all centerline segment pairs $i$ and $j$. $\mathcal{E}_{\phi}$ then applies a sequence of $N_E$ factorized attention blocks over the $N_o + N_l$ embedded scene elements, where each factorized attention block comprises a lane-to-lane, lane-to-object, and object-to-object multi-head attention layer \citep{girgis2021autobots, ngiam2022scenetransformer, liang2020lanegcn}. Lane-to-lane attention captures spatial relations between centerline segments, where the embedded lane connectivities are additionally fused into the attention keys and values. Lane-to-object attention incorporates map context into object embeddings, while object-to-object attention captures spatial relationships among objects. Following the $N_E$ factorized attention blocks, the encoder maps each object and lane embedding to latent dimensions $K_o$ and $K_l$ respectively, with both mean and variance parameterized as in a VAE \citep{kingma2014vae}. Importantly, we design $\mathcal{E}_{\phi}$ such that lane latents do not depend on the object features (\textit{i.e.,} no object-to-lane attention), to allow lane-conditioned object generation at inference. 

\textbf{Decoder} The decoder $\mathcal{D}{\gamma}$ samples from the latent distribution parameterized by the encoder  $\{ \{\mathbf{h}^{\mathcal{O}}_i\}_{i=1}^{N_o},  \{\mathbf{h}^{\mathcal{L}}_i\}_{i=1}^{N_l} \} \sim \mathcal{E}{\phi}$ and processes the embedded lane and object latents through a sequence of $N_D$ factorized attention blocks. Following these blocks, the decoder reconstructs the continuous lane and object vector inputs, supervised by an $\ell_2$ loss. For lane connectivity prediction, each pair of lane embeddings is concatenated and passed through an MLP to predict a categorical distribution over the connectivity type, trained with a cross-entropy loss. The Scenario Dreamer autoencoder is trained with the standard Evidence Lower Bound (ELBO) objective with low-$\beta$ regularization, as detailed in the Appendix.

\subsubsection{Latent Diffusion Sampling} 

We train a latent diffusion model to sample from the autoencoder's latent distribution $p(H)$ factorized as
\begin{align*}
    p(H) = \sum_{N_o, N_l} p(\{\mathbf{h}^{\mathcal{O}}_i\}_{i=1}^{N_o},  \{\mathbf{h}^{\mathcal{L}}_i\}_{i=1}^{N_l} | N_o, N_l)p(N_o, N_l).
\end{align*}
$p(N_o, N_l)$ is approximated with training set statistics, and thus we approximate $p(H)$ as $p(H) \approx p_{\theta}(H) = \sum_{N_o, N_l} p_{\theta}(\{ \{\mathbf{h}^{\mathcal{O}}_i\}_{i=1}^{N_o},  \{\mathbf{h}^{\mathcal{L}}_i\}_{i=1}^{N_l} \} | N_o, N_l) p(N_o, N_l)$. Here, the conditional distribution $p_{\theta}(\cdot|N_o, N_l)$ is parameterized with a diffusion model with weights $\theta$, which samples $N_o$ object latents and $N_l$ lane latents. Unlike image-based models, $p_{\theta}$ must accommodate variable latent sizes, so we design a customized transformer architecture with AdaLN-Zero conditioning \citep{Peebles2022DiT}. This architecture, similar to the autoencoder, is composed of a sequence of $N_{\text{DM}}$ \textit{factorized} attention blocks (illustrated in Figure \ref{fig:method}) that include sequential object-to-lane, lane-to-lane, lane-to-object, and object-to-object attention layers. This factorized processing approach allows each layer to model the layer-specific interactions while allowing for different hidden dimensions for object and lane tokens. Notably, lane tokens require a larger hidden dimension due to the high level of spatial reasoning and detail required for realistic lane generation, while object tokens are effectively represented with a smaller dimension, making the diffusion model more efficient overall. We employ the standard DDPM objective to train $p_{\theta}$, where $p_{\theta}$ is parameterized as a noise-prediction network $\epsilon_{\theta}$ that learns to predict the noise of noised lane and object latents at varying noise levels:
\begin{align*}
    L_{\text{dm}} &= \mathbb{E}_{\mathbf{H}_t, \boldsymbol{\epsilon}_t \sim \mathcal{N}(0, 1), t} \Big[ \parallel \boldsymbol{\epsilon}_{t} - \epsilon_{\theta}(\mathbf{H}_t, t) \parallel^2_2 \Big],
\end{align*}
where $\mathbf{H}_t$ denotes a stacked tensor of $N_o$ object latents and $N_l$ lane latents that is noised over $t$ forward diffusion steps with noise vector $\boldsymbol{\epsilon}_t := (\boldsymbol{\epsilon}_{\mathcal{L}}, \boldsymbol{\epsilon}_{\mathcal{O}})$ containing both the object and lane noise vectors. We refer readers to the Appendix for more details.

\textbf{Permutation Ambiguity} Unlike images, which have a natural grid-like structure, set-structured data in Scenario Dreamer poses unique challenges for diffusion models. Specifically, a phenomenon called \textit{permutation ambiguity} \citep{chen2023polydiffuse} arises: the vector elements that are sufficiently noised during training lose enough underlying structure that the model cannot reliably infer the permutation of the ground-truth signal toward which it should regress. In image-based transformer diffusion methods, such as DiT, this issue is resolved by applying a positional encoding to each grid patch. To similarly address permutation ambiguity in our model, we introduce sinusoidal positional encodings to the latent tokens prior to the factorized attention blocks.

Our positional encoding scheme involves defining an \textit{ordering} of the tokens (per token type) during training, allowing the model to better infer a noised token's likely relative spatial position and thus the ground-truth signal toward which it should regress. To achieve this, we propose a recursive ordering procedure (see Appendix): tokens are ordered by minimum $x$-value, and if $x$-values differ by less than $\epsilon$ meters, they are subsequently ordered by minimum $y$-value, then maximum $x$-value, and finally maximum $y$-value. Notably, the initial ordering by $x$-value facilitates inpainting, as tokens in $\mathcal{I}_{F_N}$ form a contiguous subsequence of ordered tokens that precede those in $\mathcal{I}_{F_P}$.

\subsubsection{Scene Generation} 

The Scenario Dreamer latent diffusion model supports multiple modes of scene generation within a single trained model: \textit{initial scene generation} to sample 64m$\times$64m scenes from $p(\mathcal{I}_F)$; \textit{lane-conditioned object generation} to sample object bounding boxes $\mathcal{O}$ conditioned on a known map $\mathcal{M}$; and \textit{scene inpainting} to sample from $p(\mathcal{I}_{F_P} | \mathcal{I}_{F_N})$. We describe each generation mode in detail in the following.

\textbf{Initial Scene Generation.} Scenario Dreamer generates novel initial driving scenes $\mathcal{I}_F$ by first sampling $(N_o, N_l) \sim p(N_o, N_l)$ from the joint distribution found in the training data, with dataset-dependent limits on $N_o$ and $N_l$ to ensure realistic scene density. Alternatively, users may directly specify $(N_o, N_l)$, which enables control over the scene density. After setting $(N_o, N_l)$, we sample latents ${ \{\mathbf{h}^{\mathcal{O}}_i\}_{i=1}^{N_o}, \{\mathbf{h}^{\mathcal{L}}_i\}_{i=1}^{N_l} }$ from the diffusion model over $T=100$ diffusion steps and decode the latents into vectorized scene elements using the autoencoder decoder.

\textbf{Lane-conditioned Object Generation} involves encoding the vectorized elements of a given map $\mathcal{M}$ using $\mathcal{E}_{\phi}$, then sampling $N_o \sim p(N_o | N_l)$ object latents from the diffusion model, which are diffused while conditioning on the encoded map latents at each denoising timestep. The resulting object latents can be decoded into object bounding box configurations with $D_{\gamma}$.

\textbf{Scene Inpainting.} Following \citep{chitta2024sledge}, we frame conditional sampling from $P(\mathcal{I}_{F_P} | \mathcal{I}_{F_N})$ as an inpainting task. Unlike grid-structured images, where boundaries between regions such as $\mathcal{I}_{F_P}$ and $\mathcal{I}_{F_N}$ are naturally defined (e.g., along $x = 0$), lane vectors in our setting can cross this boundary without restriction. To address this, we preprocess our dataset into two scene types: \textit{partitioned} scenes, which are artificially split at $x=0$, and \textit{non-partitioned} scenes, where lane vectors may span across $x=0$. The autoencoder is trained to reconstruct both scene types, while the diffusion model generates both by using a conditioning label to distinguish between them. To further enhance inpainting quality, during training the diffusion model conditions on the encoded latents of $\mathcal{I}_{F_N}$ for partitioned scenes, thus training it explicitly to inpaint. This significantly improves inpainting performance, as the model can learn to utilize the relevant scene context from $\mathcal{I}_{F_N}$ to ensure that the lane geometries remain spatially consistent across the $x=0$ boundary.

Scenario Dreamer requires specifying the number of new lane and object vectors to occupy the new 32x64 region $\mathcal{I}_{F_P}$, effectively sampling from $p(N_o^{F_P}, N_l^{F_P} | \mathcal{I}_{F_N})$. Although approximate sampling from $p(N_o^{F_P}, N_l^{F_P} | N_o^{F_N}, N_l^{F_N})$ is possible through training statistics, $p(N_l^{F_P} | \mathcal{I^{F_N}})$ is highly geometry-dependent. To account for this, we train a classifier, $f_{\phi}(N_l^{F_P} | \mathcal{M}_{F_N})$, alongside $\mathcal{E}_{\phi}$ to predict the number of lanes in $\mathcal{I}_{F_P}$ based on the context $\mathcal{I}_{F_N}$, trained only on partitioned scenes. Specifically, a learnable query vector cross-attends with the lane tokens in $\mathcal{I}_{F_N}$ within each factorized attention block of $\mathcal{E}_{\phi}$, outputting a categorical distribution over $N_o^{F_P}$ trained with a cross-entropy loss. At inference, we first sample $N_l^{F_P} \sim f_{\phi}$, followed by sampling $N_o^{F_P} \sim p(N_o^{F_P} | N_o^{F_N}, N_l^{F_N} + N_l^{F_P})$. With $(N_o^{F_P}, N_l^{F_P})$ sampled, we encode the latents for $\mathcal{I}_{F_N}$, and we generate $N_o^{F_P}$ and $N_l^{F_P}$ new tokens initialized to Gaussian noise. Standard diffusion inpainting \citep{chitta2024sledge} is then applied, where noised tokens in $\mathcal{I}_{F_N}$ are set to their encoded latents at each denoising step. The resulting latents are decoded to produce the new scene elements in $\mathcal{I}_{F_P}$.

\subsection{Behaviour Simulation}

Starting with an initial scene $\mathcal{I}$ generated by the latent diffusion model, we extend CtRL-Sim \citep{rowe2024ctrlsim}, an autoregressive Transformer-based behaviour model, to control multiple agent types (e.g., vehicles, pedestrians, and cyclists). 
To support multiple agents types, we use the $k$-disks tokenization scheme \citet{philion2024trajeglish}.
CtRL-Sim is a return-conditioned multi-agent policy designed for behaviour simulation. 
We adapt the CtRL-Sim architecture, which parameterizes the joint distribution over future returns $\returnset_t$ and actions $\jointactionset_t$, decomposed as $ p_\theta (\jointactionset_t, \returnset_t | \jointstateset_t) = \pi_\theta (\jointactionset_t | \jointstateset_t, \returnset_t) p_\theta (\returnset_t | \jointstateset_t) $. A key advantage of this decomposition is its return-conditioning, which enables exponential tilting \citep{lee2022multigamedt} of the learned return model at inference to generate good or adversarial driving behaviours. 
In the Scenario Dreamer framework, we aim to create adversarial scenarios that specifically challenge the autonomous vehicle (AV) planner. To this end, we design a reward function that penalizes collisions with the ego vehicle,
and model the discounted return $G_t = \sum_{t=t}^{t+H}r_t$ based on the cumulative rewards over a horizon $H=2$s, which we found offers improved controllability over modeling the full return.

\subsection{Simulation Framework}

With Scenario Dreamer in hand, we describe our proposed simulation framework's unique properties.
First, Scenario Dreamer supports the evaluation of AV planners within its generative simulation environments over arbitrarily long simulation lengths.
We define a route for the AV planner to follow, and simulate the other agents using CtRL-Sim.
We follow the Waymo dataset filtering scheme employed in Nocturne \citep{vinitsky2022nocturne} and GPUDrive \citep{eugene2024gpudrive} to ensure that the scenarios are valid for simulation (\textit{i.e,} absence of traffic lights, which lack annotated traffic light states). To ensure that Scenario Dreamer-generated scenarios are valid for simulation, Scenario Dreamer conditions on a binary indicator during training that indicates whether the training scene passes the Nocturne filtering scheme. At inference, we utilize classifier guidance to sample from simulation-compatible scenes. The simulation framework is described in more detail in the Appendix.

%% file: sec/5_experiments.tex
\section{Experiments}


\begin{table*}[ht]
\centering
\resizebox{\textwidth}{!}{
\begin{tabular}{@{}lccccccccccc@{}}
\toprule
& & Per-Scene & Num. & & \multicolumn{1}{c}{Perceptual Quality $\downarrow$} & \multicolumn{4}{c}{Urban Planning $\downarrow$} & Route Length $\uparrow$ & Endpoint Dist. $\downarrow$ \\
\cmidrule(r){6-6} \cmidrule(r){7-10} \cmidrule(l){11-11} \cmidrule(l){12-12}
Dataset & Method & Gen. Time (s) $\downarrow$  & Parameters & GPUh & FD & Conn. & Dens. & Reach & Conve. & (m) & (m) \\
\midrule
nuPlan & SLEDGE (DiT-L) \citep{chitta2024sledge} & $0.48$ & $539$M & $96$ & $1.89$ & $2.34$ & $2.43$ & $0.70$ & $2.44$ & $35.34$\std{$8.63$} & $0.47$\std{$0.32$} \\ 
& SLEDGE (DiT-XL) \citep{chitta2024sledge} & 0.67 & $769$M & $960$ & $1.44$ & $1.67$ & $1.74$ & $0.51$ & $1.68$ & $35.83$\std{$8.35$} & $0.42$\std{$0.29$} \\ 
\cmidrule(l){2-12}
& Scenario Dreamer (B) & \boldmath{$0.09$} & $377$M & $96$ & $1.05$ & $0.28$ & $0.45$ & $0.07$ & \boldmath{$0.14$} & \boldmath{$37.04$\std{$10.21$}} & $0.32$\std{$0.80$} \\
& Scenario Dreamer (L) & $0.16$ & $679$M & $256$ & \boldmath{$0.67$} & \boldmath{$0.18$} & \boldmath{$0.43$} & \boldmath{$0.03$} & $0.33$ & $36.87$\std{$10.37$} & \boldmath{$0.25$\std{$0.71$}} \\
\midrule
\midrule
Waymo & DriveSceneGen (GT Raster)$^*$ \citep{sun2024drivescenegen} & - & - & - & $40.59$ & $4.53$ & $1.18$ & $0.64$ & $5.58$ & \boldmath{$41.61$\std{$18.61$}} & \boldmath{$0.01$\std{$0.00$}} \\
\cmidrule(l){2-12}
& Scenario Dreamer (B) & \boldmath{$0.08$} & $376$M & $96$ & $1.61$ & $0.17$ & $1.05$ & $0.56$ & $3.81$ & $38.23$\std{$12.78$} & $0.32$\std{$0.90$} \\
& Scenario Dreamer (L) & $0.16$ & $678$M & $256$ & \boldmath{$1.38$} & \boldmath{$0.03$} & \boldmath{$0.95$} & \boldmath{$0.28$} & \boldmath{$2.05$} & $38.92$\std{$13.56$} & $0.21$\std{$0.75$} \\
\bottomrule
\end{tabular}
}
\caption{Assessment of lane graph generation evaluated on the Waymo Open Motion and nuPlan test datasets. For each metric and dataset, the best method is \textbf{bolded}. $^*$ denotes privileged version of the method.} 
\label{tab:lanegraphgenerationresults}
\end{table*}

\begin{table*}[ht]
\centering
\resizebox{0.7\textwidth}{!}{
\begin{tabular}{@{}lcccccccc@{}}
\toprule
& & \multicolumn{6}{c}{Distributional JSD $\downarrow$} & Collision Rate $\downarrow$ \\ 
\cmidrule(r){3-8} \cmidrule(l){9-9} 
Dataset & Method & Near. Dist. & Lat. Dev. & Ang. Dev. & Len. & Wid. & Speed & (\%) \\ 
\midrule
nuPlan & SLEDGE (DiT-L) \citep{chitta2024sledge} & $0.53$ & $0.63$ & $3.60$ & $11.84$ & $10.16$ & $0.46$ & $22.3$ \\ 
& SLEDGE (DiT-XL) \citep{chitta2024sledge} & 0.49 & 0.49 & 3.26 & 11.16 & 10.29 & 0.47 & 21.2 \\ 
\cmidrule(l){2-9}
& Scenario Dreamer (B) & $0.12$ & $0.15$ & \boldmath{$0.17$} & \boldmath{$0.22$} & \boldmath{$0.14$} & $0.07$ & $11.9$ \\ 
& Scenario Dreamer (L) & \boldmath{$0.09$} & \boldmath{$0.11$} & $0.18$ & $0.25$ & $0.17$ & \boldmath{$0.06$} & \boldmath{$9.3$} \\ 
\midrule 
\midrule
Waymo & DriveSceneGen (GT Raster)$^*$ \citep{sun2024drivescenegen} & $0.63$ & $1.01$ & $2.43$ & $58.86$ & $54.51$ & $18.70$ & \boldmath{$0.2$} \\
\cmidrule(l){2-9}
& Scenario Dreamer (B) & $0.07$ & $0.07$ & \boldmath{$0.07$} & \boldmath{$0.40$} & \boldmath{$0.25$} & \boldmath{$0.36$} & $5.4$ \\ 
& Scenario Dreamer (L) & \boldmath{$0.06$} & \boldmath{$0.05$} & \boldmath{$0.07$} & $0.42$ & \boldmath{$0.25$} & $0.38$ & $4.8$ \\ 
\bottomrule
\end{tabular}
}
\caption{Assessment of initial agent bounding box generation evaluated on the Waymo Open Motion and nuPlan test datasets. $^*$ denotes privileged version of the method.}
\vspace{-10pt}
\label{tab:agentgenerationresults}
\end{table*}

\begin{table*}[ht]
\centering
\resizebox{\textwidth}{!}{
\begin{tabular}{@{}lccccccccc@{}}
\toprule
& Per-Scene & Perceptual Quality $\downarrow$ & \multicolumn{4}{c}{Urban Planning $\downarrow$} & Route Length $\uparrow$ & Endpoint Dist. $\downarrow$ \\
\cmidrule(r){3-3} \cmidrule(r){4-7} \cmidrule(l){8-8} \cmidrule(l){9-9}
Method & Gen. Time (s) & FD & Conn. & Dens. & Reach & Conve. & (m) & (m) \\
\midrule
Scenario Dreamer (Non-factorized) & $0.20$ & $1.21$ & $0.51$ & $0.67$ & $0.16$ & $0.55$ & $36.71$\std{$9.95$} & $0.37$\std{$0.76$} & \\
Scenario Dreamer (No lane ordering) & \boldmath{$0.09$} & $1.36$ & $0.60$ & \boldmath{$0.07$} & $0.21$ & $0.71$ & $36.70$\std{$9.94$} & $0.35$\std{$0.74$} & \\
Scenario Dreamer (Heuristic topology) & \boldmath{$0.09$} & \boldmath{$1.05$} & \boldmath{$0.28$} & $0.44$ & $0.12$ & $0.60$ & \boldmath{$37.13$\std{$9.85$}} & \boldmath{$0.30$\std{$0.23$}} & \\
\midrule
Scenario Dreamer & \boldmath{$0.09$} & \boldmath{$1.05$} & \boldmath{$0.28$} & $0.45$ & \boldmath{$0.07$} & \boldmath{$0.14$} & $37.04$\std{$10.21$} & $0.32$\std{$0.80$} & \\
\bottomrule
\end{tabular}
}
\caption{Ablation experiments validating the design choices of Scenario Dreamer (B) for lane graph generation on the nuPlan dataset.}
\vspace{-10pt}
\label{tab:ablations}
\end{table*}

\begin{table}[h]
\centering
\resizebox{\columnwidth}{!}{
\begin{tabular}{@{}clccccc@{}}
\toprule
Other Agent Beh. & Test Env. & Avg. Route Len. (m) &  Coll. (\%) & Offroad (\%) & Succ. (\%) \\
\cmidrule(l){1-3} \cmidrule(l){4-6}
Log Replay & Waymo Test & 55 & $29.3$\std{$0.6$} & $6.9$\std{$0.8$} & $63.8$\std{$0.8$} \\
CtRL-Sim (Pos. Tilt) & Waymo Test & 55 & $35.7$\std{$0.7$} & $4.9$\std{$0.9$} & $59.4$\std{$0.4$} \\
\midrule
CtRL-Sim (Pos. Tilt) & SD (55m) & 55 & $33.8$\std{$1.2$} & $6.4$\std{$0.7$} & $59.8$\std{$1.7$} \\
CtRL-Sim (Pos. Tilt) & SD (100m) & 100 & $52.8$\std{$1.4$} & $9.1$\std{$1.6$} & $38.2$\std{$1.9$} \\
CtRL-Sim (Neg. Tilt) & SD (100m) & 100 & $59.0$\std{$1.3$} & $9.0$\std{$0.8$} & $32.1$\std{$1.3$} \\
\bottomrule
\end{tabular}
}
\caption{\textbf{RL Planner Results.} PPO agents are trained in GPUDrive on 100 Waymo scenes and evaluated on 250 scenes (Waymo test scenes or 55m/100m-route Scenario Dreamer (SD) scenes). mean\std{\text{std}} reported over 5 seeds.}
\label{tab:sim_results}
\vspace{-15pt}
\end{table}

\textbf{Datasets} The Scenario Dreamer latent diffusion model is separately trained on both the Waymo Open Motion Dataset \citep{ettinger2021womd} and the nuPlan dataset \citep{caesar2024nuplan}. For Waymo, our model captures vehicles, pedestrians, and cyclists, and we process the lane centerlines and their connectivity, excluding other map elements. For nuPlan, we model vehicles, pedestrians and static objects, and we process the lane centerlines, lane connectivity, and traffic light states. For both datasets, we use a $64$m $\times 64$m FOV centered on the ego vehicle and centerlines are processed using the compression algorithm from \citet{chitta2024sledge}, detailed in the Appendix. 

\textbf{Metrics} To evaluate the quality of initial scene generations, we assess our methods on both lane graph generation and initial agent bounding box generation tasks. For lane graph generation, we report \textbf{Urban Planning} metrics, following prior works \citep{mi2021hdmapgen, sun2024drivescenegen, chitta2024sledge}. These metrics measure the distributional realism of the generated lane graph connectivity by computing Frechet distances on node features for nodes with degree $\neq 2$, referred to as \textit{key points}. The following features are computed: \textit{Connectivity} computes the degrees of all key points across the lane graphs; \textit{Density} computes the number of key points in each lane graph; \textit{Reach} computes the number of available paths from each keypoint to others; and \textit{Convenience} computes the Dijsktra path lengths for all valid paths between key points. As in \citet{chitta2024sledge}, the Urban Planning metrics are scaled by suitable powers of 10 for readability. \textbf{Frechet Distance (FD)} measures the perceptual quality of the generated spatial lane positions by computing Frechet distances between the penultimate layer lane embeddings of a separately trained Scenario Dreamer autoencoder model. We additionally measure the longest \textbf{Route Length} by traversing the generated lane graph from the origin, and \textbf{Endpoint Distance}, which measures the average distance between the endpoint of lane $i$ and starting point of lane $j$ for all predicted successor edges $(i,j)$ in the generated lane graphs. All lane graph generation metrics are computed on $50$k real lane graphs and $50$k ground-truth test lane graphs. 

For initial agent generation, we focus on the distributional realism of vehicles, reporting Jensen-Shannon Divergence (\textbf{JSD}) metrics as in \citet{lu2024scenecontrol}, including:  \textit{Nearest Dist.} between each vehicle and its neighbours, \textit{Lateral Dev.} from the closest centerlines, \textit{Angular Dev.} from the closest centerline, \textit{Length} and \textit{Width} of the vehicles, and \textit{Speed} of the vehicles. The JSD metrics are computed over $50$k real and generated scenes and scaled by suitable powers of 10. We also report the \textbf{Collision  Rate} of the agent bounding boxes. Behaviour simulation metrics are similar to that of the agent generation, with details in the Appendix.

\textbf{Methods under Comparison} We evaluate Scenario Dreamer against competitive diffusion-based methods for initial scene generation on both the nuPlan and Waymo datasets. For nuPlan, we compare with the SLEDGE DiT-L and DiT-XL models \citep{chitta2024sledge}, retraining them following their open-source code. SLEDGE encodes driving scenes into a latent space using an image rasterization scheme, which is then modeled by a latent diffusion process and decoded in a vectorized format. On the Waymo dataset, we benchmark against DriveSceneGen \citep{sun2024drivescenegen}. However, in our experiments, the DriveSceneGen diffusion model did not converge, so we instead use a privileged version that provides an \textit{upper bound} on DriveSceneGen’s performance by rasterizing $50$k ground-truth training scenes following their rasterization pipeline and decoding them with DriveSceneGen’s vectorization post-processing. Notably, Scenario Dreamer is the only model that fully processes scene elements in a vectorized manner. For behaviour simulation, we compare CtRL-Sim against the rule-based IDM \citep{treiber2000idm} and a data-driven baseline Trajeglish \citep{philion2024trajeglish}, with quantitative and qualitative results reported in the Appendix. The Scenario Dreamer latent diffusion model is available in two sizes: the base model (B) with 377M parameters, trained over 24 hours on 4 A100 GPUs, and the large model (L) with 679M parameters, trained over 32 hours on 8 A100 GPUs. Additional model details are provided in the Appendix.

\textbf{Results} Table \ref{tab:lanegraphgenerationresults} reports the lane graph generation results on the nuPlan and Waymo datasets, along with per-scene generation times, parameter counts, and diffusion model GPU training hours (GPUh) for each method. On the nuPlan dataset, Scenario Dreamer outperforms SLEDGE across all lane graph generation metrics. Notably, the smaller Scenario Dreamer base model surpasses SLEDGE's largest DiT-XL model on every metric, with approximately $2\times$ fewer parameters, $7\times$ lower inference latency, and $10\times$ fewer GPU training hours. We note that our reproduction of the SLEDGE DiT-XL model underperformed the results reported in \citet{chitta2024sledge}. Nevertheless, Scenario Dreamer (L) still outperforms the  published results in \citet{chitta2024sledge} across all metrics. These results highlight the effectiveness of the vectorized design for lane graph generation. 

The vectorized processing enhances efficiency by focusing parameters directly on the relevant vectorized scene elements rather than processing empty pixels and is not sensitive to the specific rasterization scheme that is employed. With vectorized processing, each latent corresponds to a scene element, allowing the model to effectively capture and learn the interactions between scene elements. Practically, this results in Scenario Dreamer's latent diffusion model being substantially more efficient than SLEDGE, as the vectorized latent representation requires far fewer Transformer layers to model effectively. Concretely, the largest SLEDGE diffusion model requires 28 Transformer layers for processing, whereas Scenario Dreamer's largest latent diffusion model requires only $6$ lane to lane attention layers. On the Waymo dataset, Scenario Dreamer outperforms the privileged version of the DriveSceneGen model on the Urban Planning and perceptual metrics. This advantage is attributed to substantial errors introduced by vectorizing a rasterized scene as a postprocessing step as done by DriveSceneGen, underscoring the importance of directly processing vectorized elements within the model itself. 

Table \ref{tab:agentgenerationresults} presents the results on agent generation on the nuPlan and Waymo datasets. Consistent with the lane graph generation results, the Scenario Dreamer model achieves more realistic agent configurations, as reflected by lower JSD metrics and fewer collisions compared to related methods. In the Appendix, we visualize a random sample of Scenario Dreamer, SLEDGE, and privileged DriveSceneGen generated scenes, where Scenario Dreamer scenes exhibit visually more realistic lane graph and agent configurations. We attribute these improvements to the model’s vectorized scene processing approach.

Table \ref{tab:ablations} reports ablation experiments examining the impact of key architectural design choices. First, when we replace the learned lane connectivity prediction task with heuristic labels based on lane endpoint distances and relative orientations (\textit{Heuristic topology}), Scenario Dreamer shows degraded performance on the Reach and Connectivity Urban Planning metrics, which best reflect the realism of lane graph connectivity. This result indicates that learning lane connectivity directly outperforms post-processing heuristics. Additionally, removing the proposed lane ordering (\textit{No lane ordering}) introduces permutation ambiguity, which reduces the generation quality. Finally, we trained a non-factorized Scenario Dreamer diffusion model (\textit{Non-factorized}) with a comparable parameter count and observed that factorized processing not only enhances performance but also reduces inference latency by approximately $2\times$, underscoring the effectiveness of the factorized design.


Table \ref{tab:sim_results} summarizes the performance of a PPO \citep{schulman2017proximal} planner trained in GPUDrive \citep{eugene2024gpudrive} on 100 Waymo training scenes, reformatted to be compatible with Scenario Dreamer simulation environments. We evaluate this planner in both generative Scenario Dreamer environments with 55m \textit{(SD (55m))} and 100m (\textit{SD (100m))} routes and in non-generative Waymo test environments \textit{(Waymo Test)}. For consistency, the Waymo maps are processed to retain only centerlines, ignoring other map elements. We measure collision rate (\textit{Coll.}) with other agents, offroad rate (\textit{Offroad}) based on a lateral deviation of more than 2.5m from the route, and success rate (\textit{Succ.}) determined by whether the planner completes the route. The results indicate that replacing non-reactive log replay agent behaviors with reactive CtRL-Sim behaviors has only a marginal impact on the RL policy’s performance (Row 1 vs. Row 2), underscoring the realism of CtRL-Sim. Moreover, evaluating the RL agents on Scenario Dreamer-generated environments with 55m routes—matching the average route length of the Waymo test scenes—yields comparable planner performance (Row 2 vs. Row 3), further affirming the fidelity of Scenario Dreamer. We show that increasing the route length to 100m notably degrades policy performance (Row 3 vs. Row 4), and negative tilting of CtRL-Sim further exacerbates this effect (Row 4 vs. Row 5), highlighting how longer and more adversarial scenarios challenge the RL planner.

%% file: sec/6_conclusion.tex
\section{Conclusion}

We present Scenario Dreamer, a fully data-driven generative simulator for autonomous driving planners.
At its core, Scenario Dreamer is comprised of an initial scene generator and a return-conditioned multi-agent Transformer behaviour model.
A key novelty of this work is the vectorized latent diffusion model, enabling efficient and effective scene generation compared to rasterized scene encoding approaches. We hope that the Scenario Dreamer framework can be the foundation of future research on fully data-driven generative simulators for autonomous driving research and development.

\textbf{Limitations} We observed qualitatively that the  traffic light signaling of lanes did not always provide valid traffic logic. Furthermore, Scenario Dreamer currently only generates centerline maps. In the future, we plan to generate other road element types (e.g., road edges, crosswalks).

%% file: supplemental/results.tex
\subsection{Behaviour Simulation}

\begin{table}[h]
\centering
\resizebox{\columnwidth}{!}{
\begin{tabular}{@{}lccccccc@{}}
\toprule
& & \multicolumn{4}{c}{JSD ($\times 10^{-2}$) $\downarrow$} & Agent Collision & Planner Collision \\
\cmidrule(l){3-6} \cmidrule(l){7-8}
Method & Control? & Lin. Spd. & Ang. Spd. & Acc. & Near. Dist. & (\%) $\downarrow$ & (\%) $\downarrow$ \\
\midrule
IDM \citep{treiber2000idm} & \ding{55} & $9.2$ & $0.4$ & $19.8$ & $1.6$ & $7.2$ & $5.8$ \\
Trajeglish$^\dagger$ \citep{philion2024trajeglish} & \ding{55} & $19.5$ & $0.3$ & \boldmath{$19.7$} & $4.0$ & $6.4$ & $7.0$ \\
\midrule
CtRL-Sim (Positive Tilting) & \ding{51} & \boldmath{$4.1$} & \boldmath{$0.1$} & $20.1$ & \boldmath{$1.3$} & \boldmath{$6.2$} & \boldmath{$4.9$} \\ 
CtRL-Sim (Negative Tilting) & \ding{51} & $4.2$ & $0.2$ & $26.1$ & $1.5$ & $10.9$ & $11.9$ \\ 
\bottomrule
\end{tabular}
}
\caption{Comparison of different methods for behaviour simulation over 1000 Waymo test scenes. $^\dagger$ indicates reimplementation. }
\label{tab:behaviour_sim_results}
\end{table}

Table \ref{tab:behaviour_sim_results} presents the behaviour simulation results, comparing various methods on a held-out test set of simulation-compatible Waymo scenes. We evaluate against the rule-based baseline IDM \citep{treiber2000idm}, which is employed as the behaviour model in SLEDGE \citep{chitta2024sledge}, as well as the competitive data-driven behaviour model Trajeglish \citep{philion2024trajeglish}. All methods operate on the Scenario Dreamer lane graph representation, where each Waymo map is processed as a set of lane centerlines resampled to 50 points per lane segment and a corresponding lane graph. The IDM policy assigns a random route for each agent to follow by traversing the lane graph. Simulations are performed within an $80\text{m} \times 80 \text{m}$ field of view (FOV) around the ego vehicle at each timestep. The ego vehicle uses an IDM planner that follows the lane centerline route closest to the logged trajectory in the dataset. We report standard JSD distributional realism metrics alongside collision rates: (\textit{Agent Collision}), the rate of collisions among simulated agents, and (\textit{Planner Collision}), the rate of collisions between simulated agents and the IDM planner. To showcase the controllability of CtRL-Sim, we include results for models with exponentially tilted behaviour: $\kappa = +10$ (\textit{CtRL-Sim Positive Tilting}) and $\kappa = -50$ (\textit{CtRL-Sim Negative Tilting}). 

We observe that the CtRL-Sim behaviour model performs competitively with both data-driven and rule-based baselines across JSD and collision rate metrics. Notably, the positively-tilted CtRL-Sim model surpasses the baselines in all JSD metrics except acceleration JSD, while achieving the lowest planner collision rate ($4.9\%$) due to the positive tilting. By contrast, the IDM and Trajeglish behaviour models exhibit higher planner collision rates, demonstrating their inability to effectively coordinate with the IDM planner. A key advantage of CtRL-Sim is its ability to intuitively control adversarial behaviours. Even with negative tilting, CtRL-Sim maintains realistic driving behaviours, evidenced by only a modest increase in JSD metrics, while the planner collision rate rises by $7$ percentage points. Importantly, this version of CtRL-Sim has not been fine-tuned on adversarial driving data, unlike prior work in \citep{rowe2024ctrlsim}.

\subsection{Qualitative Results}

\begin{figure*}[tbp]
  \centering
  \includegraphics[width=0.9\textwidth]{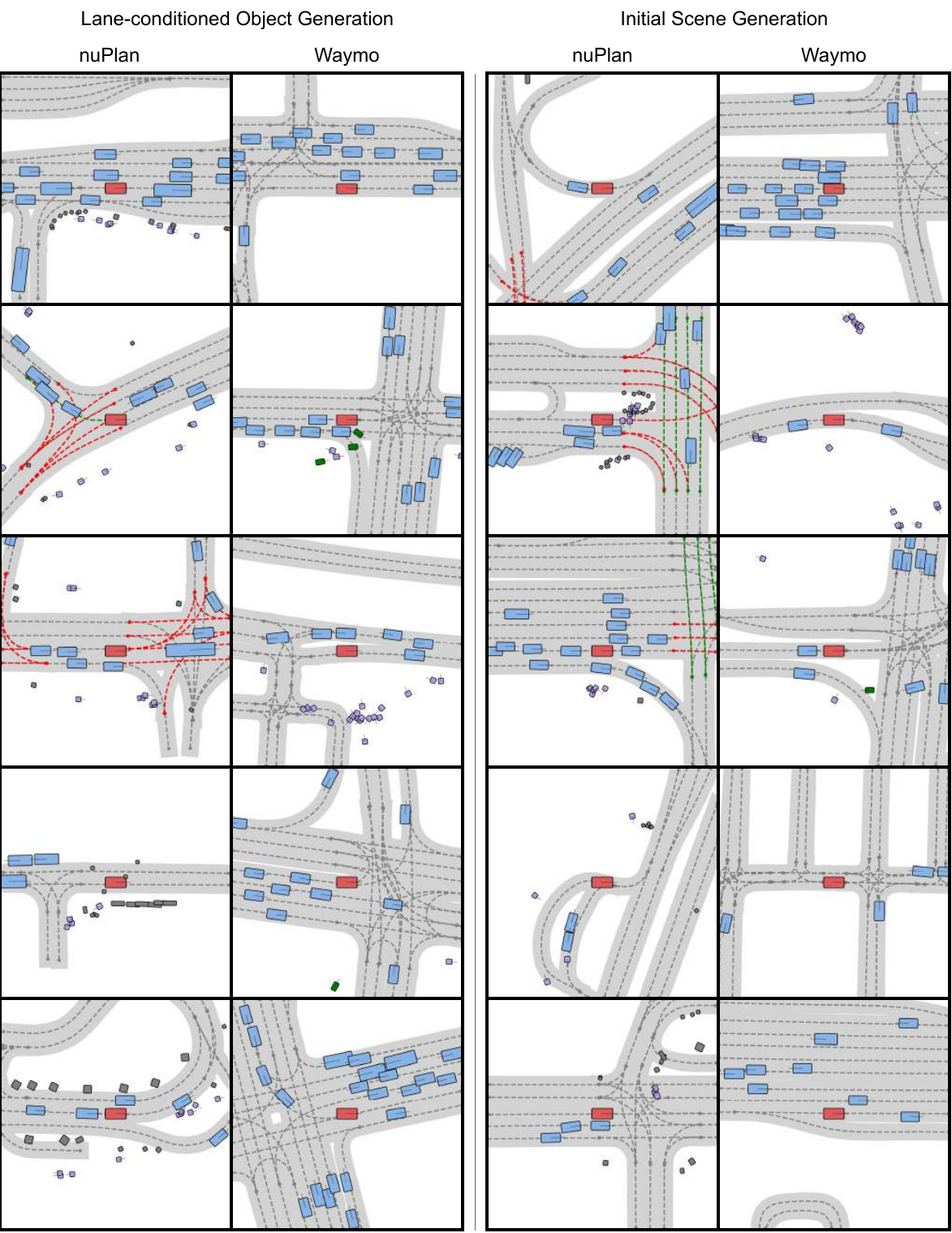}
  \caption{\textbf{Lane-conditioned object generation and initial scene generation qualitative results.} We visualize samples from the Scenario Dreamer (L) model in lane-conditioned object generation mode (columns 1 and 2) and full scene generation mode (columns 3 and 4) on both the nuPlan and Waymo datasets.}
  \label{fig:generationmodes}
\end{figure*}

\begin{figure*}[tbp]
  \centering
  \includegraphics[width=0.9\textwidth]{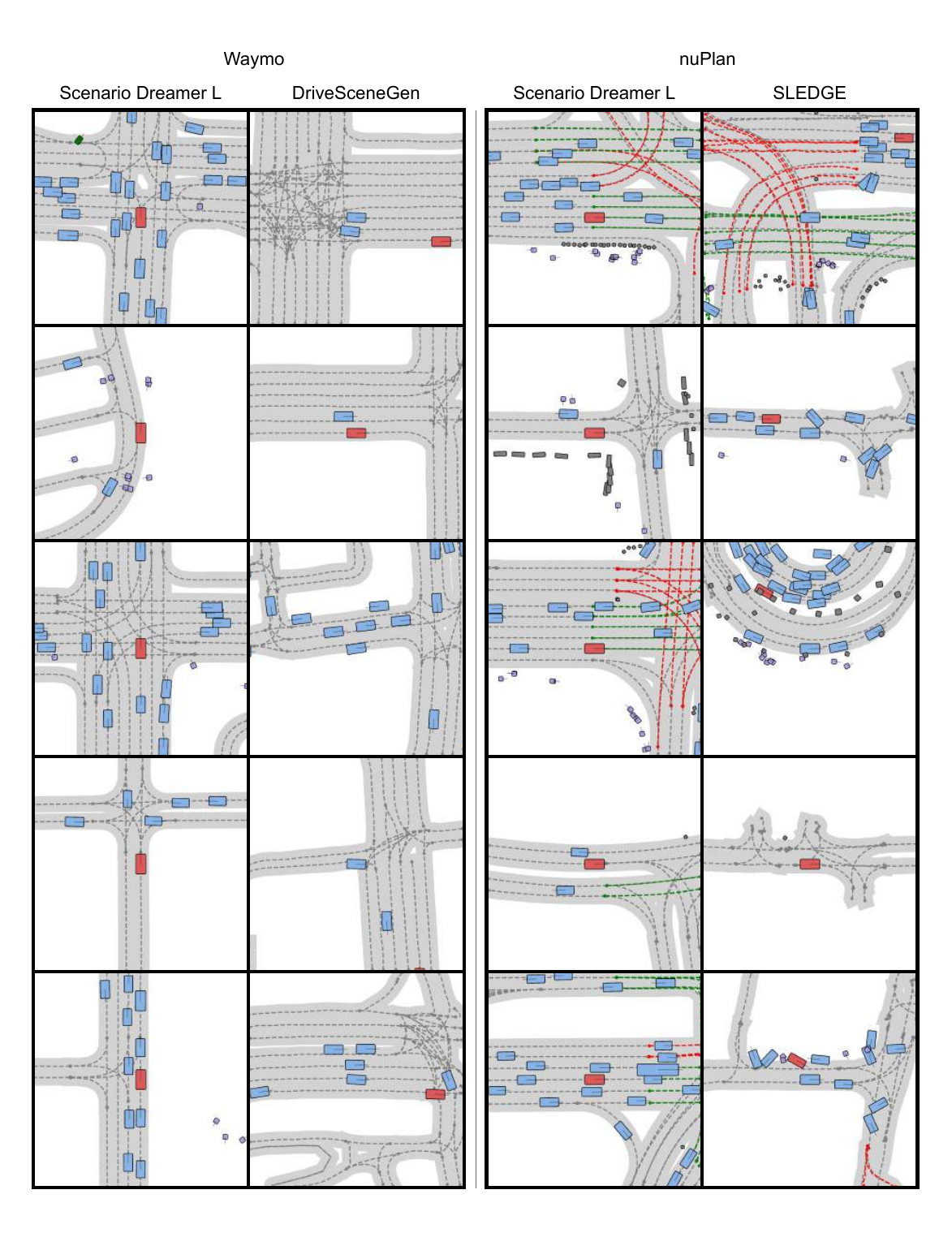}
  \caption{\textbf{Qualitative results comparison across methods (set 1).} We visualize samples from the Scenario Dreamer (L) model trained on the Waymo dataset and the privileged DriveSceneGen model (columns 1 and 2) along with samples from Scenario Dreamer (L) and SLEDGE DiT-XL both trained on nuPlan (columns 3 and 4).}
  \label{fig:methods_examples1}
\end{figure*}

\begin{figure*}[tbp]
  \centering
  \includegraphics[width=0.9\textwidth]{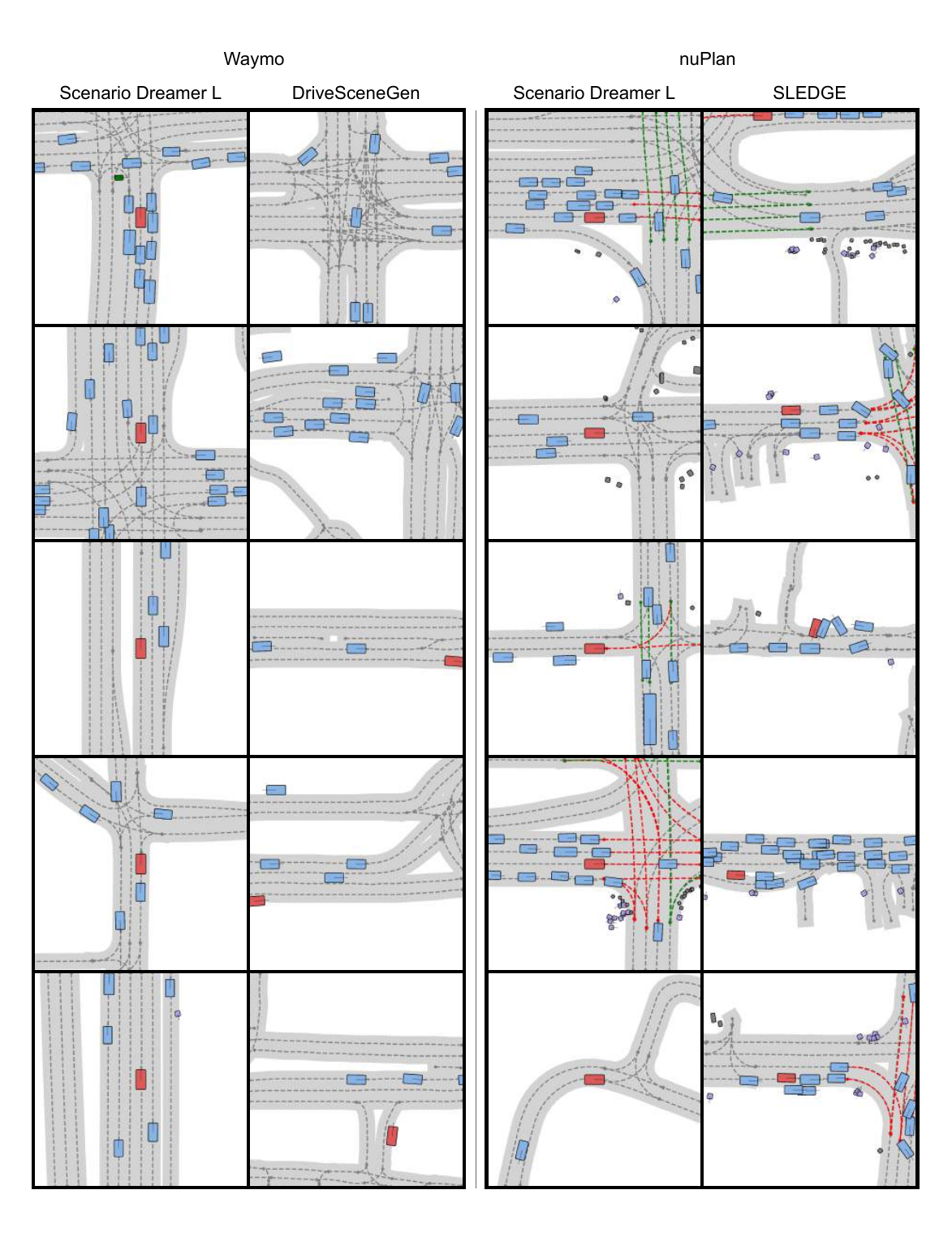}
  \caption{\textbf{Qualitative results comparison across methods (set 2).} We visualize samples from the Scenario Dreamer (L) model trained on the Waymo dataset and the privileged DriveSceneGen model (columns 1 and 2) along with samples from Scenario Dreamer (L) and SLEDGE DiT-XL both trained on nuPlan (columns 3 and 4).}
  \label{fig:methods_examples2}
\end{figure*}

Figure \ref{fig:generationmodes} showcases samples generated by the Scenario Dreamer L model in \textit{lane-conditioned object generation} and \textit{initial scene generation} modes, trained on both the nuPlan and Waymo datasets. Figures \ref{fig:methods_examples1} and \ref{fig:methods_examples2} compare random initial scene samples from our Scenario Dreamer L model (trained on nuPlan) with those from the retrained SLEDGE DiT-XL model \citep{chitta2024sledge}. Despite having $100$M fewer parameters, $4\times$ lower inference latency, and $\sim4\times$ fewer GPU training hours, Scenario Dreamer L produces higher-quality scenes. Additionally, Figures \ref{fig:methods_examples1} and \ref{fig:methods_examples2} compares rasterized ground-truth samples processed through the DriveSceneGen vectorization pipeline \citep{sun2024drivescenegen} with random samples from Scenario Dreamer L trained on the Waymo dataset. The vectorization pipeline introduces notable errors in lane graph reconstruction, often resulting in implausible configurations. In contrast, Scenario Dreamer L generates significantly more realistic lane graphs, demonstrating the effectiveness of the proposed vectorized approach. 

\begin{figure*}[tbp]
  \centering
  \includegraphics[width=0.9\textwidth]{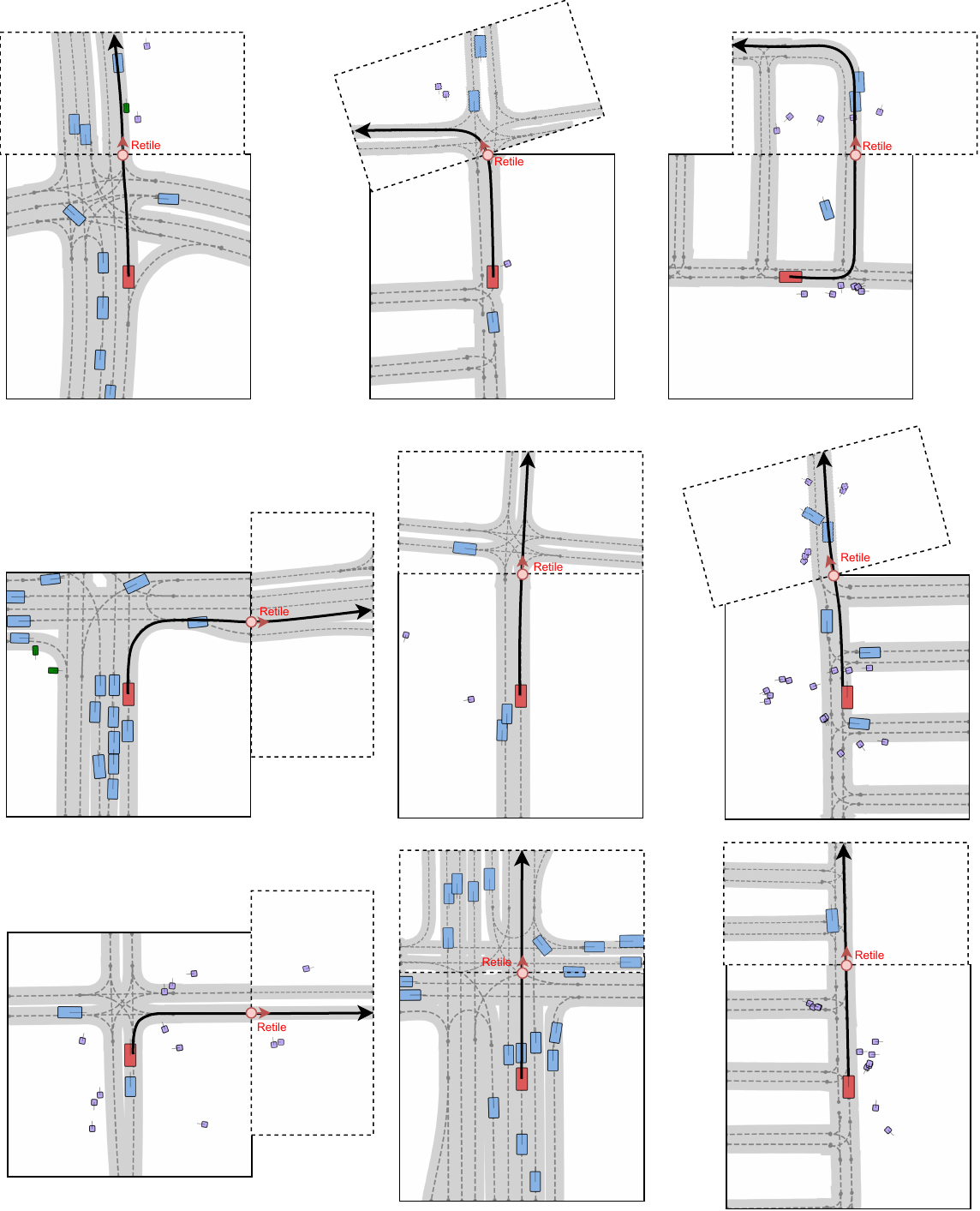}
  \caption{\textbf{Scenario Dreamer inpainting.} We visualize the inpainting capabilities of Scenario Dreamer trained on the Waymo dataset. The initial tile is outlined in a solid line, and the new inpainted tile is outlined in a dashed line. The ego route is visualized as a solid black line.}
  \label{fig:inpainting-large}
\end{figure*}

\begin{figure*}[tbp]
  \centering
  \includegraphics[width=0.9\textwidth]{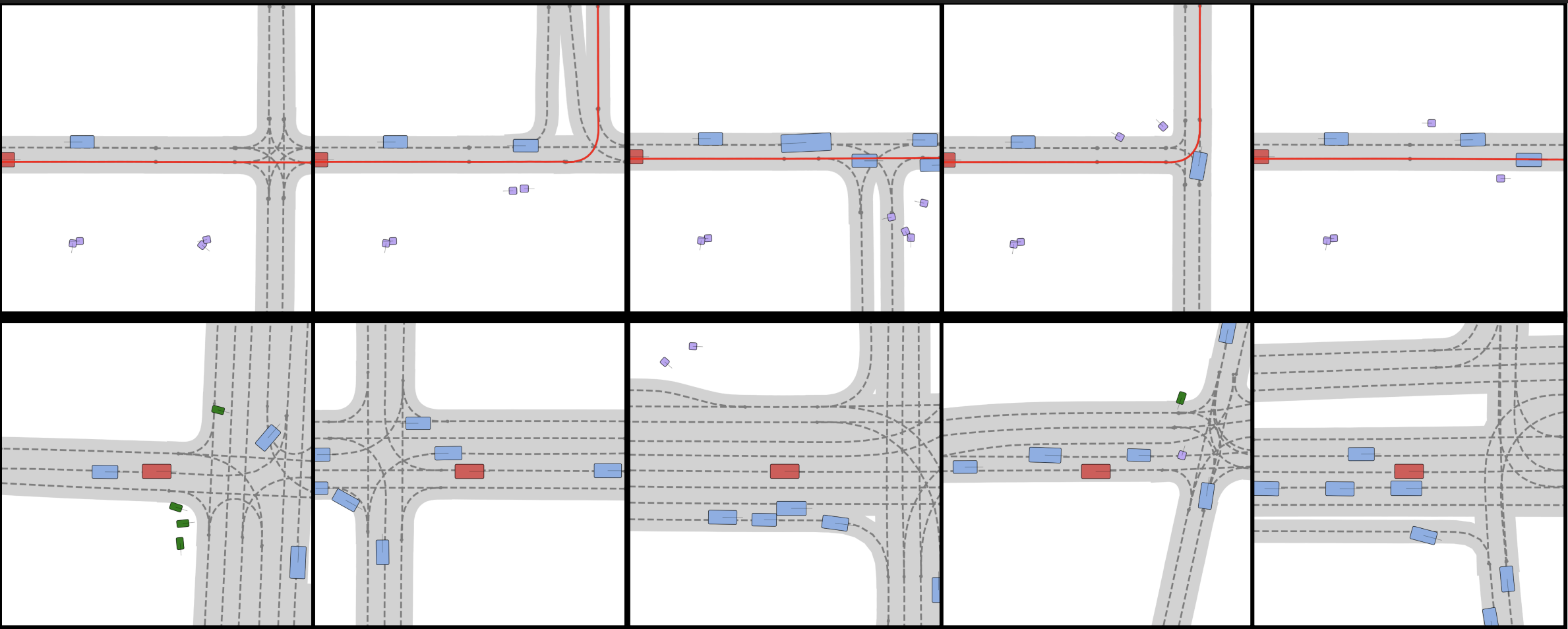}
    \caption{\textbf{Scene Diversity}: (Top) Right-half of the scene inpainted from the same left-half of the scene. (Bottom) Random samples with the same initial conditions: 8 agents and 24 lanes.}
    \label{fig:diversity}
\end{figure*}

\begin{figure*}[tbp]
  \centering
  \includegraphics[width=0.9\textwidth]{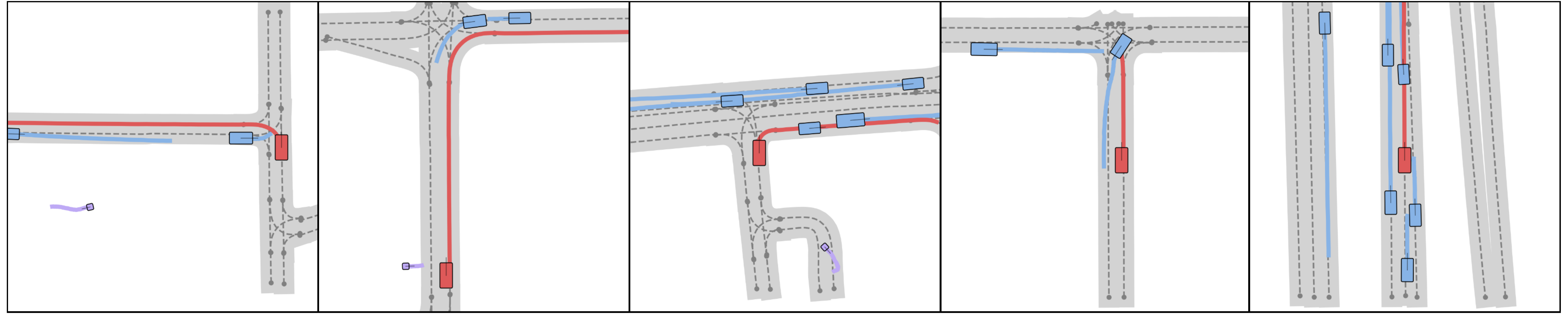}
    \caption{\textbf{Behaviour Simulation Examples}: CtRL-Sim \color{custom_blue}rollouts \color{black} with IDM \color{custom_red}ego planner \color{black} simulated from SD initial scenes.}
    \label{fig:behaviour}
\end{figure*}

Figure \ref{fig:inpainting-large} illustrates the inpainting capabilities of the Scenario Dreamer L model trained on the Waymo dataset. Scenario Dreamer is able to inpaint at the border of complex lane geometries, such as intersections. We refer readers to the Scenario Dreamer webpage, where we visualize the diffusion chain of Scenario Dreamer for the different supported generation modes. Figure \ref{fig:diversity} showcases the diversity of generated Scenario Dreamer scene generations by visualizing multiple randomly generated scenes under the same initial conditions: (Top) inpainting on the same partial scene and (Bottom) full scene generation conditioned on the same number of lanes (24) and agents (8). In both cases, the generated scenes demonstrate plausible diversity. Figure \ref{fig:behaviour} shows qualitative examples of closed-loop CtRL-Sim behaviour model rollouts simulated from Scenario Dreamer generative environments.

\FloatBarrier

%% file: supplemental/simulation.tex
\subsection{Simulator Properties}

Scenario Dreamer supports the evaluation of AV planners within its generative simulation environments over arbitrarily long simulation lengths. Specifically, it makes it possible to define a $500$-meter route for the AV planner to follow. To construct such a simulation environment, an initial scene $\mathcal{I}_F$ is generated from the initial scene generator, and a route is selected. Then, an SE(2) transformation is applied to the generated scene $\mathcal{I}_F$ such that $\mathcal{I}_F$ is renormalized to the end of the driving route in $\mathcal{I}_F$. The Scenario Dreamer diffusion model subsequently inpaints $\mathcal{I}_{F_P}$ based on the existing map $\mathcal{I}_{F_N}$, iterating this process of route selection and inpainting until the desired route length is achieved. Routes are represented by lane centerlines, resampled at $1$-meter intervals, and determined by a depth-first traversal of the lane graph starting from the origin.

The Scenario Dreamer model trained on the Waymo dataset serves as the foundation for deriving simulation environments used for AV planner evaluation. Maps are represented using the compact centerline-based format proposed in SLEDGE \citep{chitta2024sledge}. Each lane segment consists of 50 points, along with lane connectivity information. The current version of Scenario Dreamer does not include additional map elements such as crosswalks, stop signs, and road edges; however, we plan to incorporate these elements in future work. The simulator supports vehicle, pedestrian, and bicyclist agent types. Agents are controlled using the trained CtRL-Sim model, which operates in the \(\Delta x, \Delta y, \Delta \theta\) action space with actions selected from the \(k\)-disks vocabulary, visualized in Figure \ref{fig:kdisks}. The AV planner outputs steering and acceleration commands, which are passed through a forward bicycle model \citep{eugene2024gpudrive} to update the AV’s state, ensuring physical realism in the trajectory. The AV planner's context is defined by a $64$m radius around the planner's position at each timestep. The AV planner’s objective is to complete the route within a reasonable time limit without colliding or deviating from the path by more than $2.5$ meters. The simulation is deemed a failure if the AV collides with an agent, exceeds the deviation threshold, or surpasses the time limit.



\subsection{Fully Data-Driven Simulation}

Unlike \citet{chitta2024sledge}, which employ a rule-based IDM model for behaviour simulation, Scenario Dreamer leverages a data-driven and controllable CtRL-Sim behaviour model, offering enhanced scenario diversity and realism. The AV planner operates within a simulation radius of $64$m centered on the AV's current position. To accommodate this, CtRL-Sim simulates agents within a slightly larger $80$m$\times80$m field of view (FOV) around the AV’s position. This restricted FOV minimizes simulation latency, while agents outside the radius remain in a constant state until they enter the FOV. To ensure smooth transitions, agents that exit the FOV are not allowed to reenter. By limiting behaviour modeling to this restricted FOV, we achieve efficiency using a lightweight 8M parameter behaviour model—at least 
$5\times$ smaller than several state-of-the-art behaviour models trained on the Waymo dataset \citep{philion2024trajeglish, wang2023mvte, hu2024gump}.

\subsection{Adversarial Simulation Environments}

Scenario Dreamer supports the generation of adversarial simulation environments by leveraging the controllability of CtRL-Sim through exponential tilting. In an adversarial simulation environment, we ensure at least one agent within the simulation radius is sampling from a negatively tilted return distribution, which encourages adversarial collision behaviours with the ego vehicle. We show systematically that these adversarial tilting produces more collisions with an IDM planner in Table \ref{tab:behaviour_sim_results}.

\subsection{Training Scenario Dreamer-compatible RL Agents in GPUDrive}


We train PPO \citep{schulman2017proximal} planning agents on 100 simulation-compatible Waymo scenarios for 100 million steps using the GPU-accelerated GPUDrive simulator \citep{eugene2024gpudrive}. Training requires approximately 24 hours on a single A100-L GPU. To ensure compatibility with Scenario Dreamer environments, we make four key modifications to the GPUDrive simulator:

\begin{enumerate}
    \item \textbf{Route-conditioning}: We replace goal-conditioning with route-conditioning, as Scenario Dreamer provides a route for the planner to follow. The route is defined by the sequence of lane centerlines nearest to the logged ground-truth trajectory of the ego.
    \item \textbf{Map simplification}: We remove road edges and crosswalks from the map, retaining only lane centerlines that are preprocessed in the same way as in Scenario Dreamer.
    \item \textbf{Route deviation penalty}: To encourage on-road driving in the absence of road edges, we penalize the planner for lateral deviations exceeding $2.5 + \text{width} = 4.83\text{m}$ from the lane centerline.
    \item \textbf{Episode termination}: Episodes end only when the ego either reaches the end of the prescribed route (i.e., the goal) or when the episode times out after 9 seconds.
\end{enumerate}

The reward function consists of three components: a collision penalty with weight 0.1, a route-deviation penalty with weight 0.05, and a goal-achievement reward with weight 1.

%% file: supplemental/model.tex
\subsection{Vectorized Latent Diffusion Model}

\subsubsection{Autoencoder}

\textbf{Encoder} The autoencoder encoder $\mathcal{E}{\phi}$ takes as input a set of $N_l$ lanes $\{ \mathbf{l}_i \}_{i=1}^{N_l}$, $N_o$ objects $\{ \mathbf{o}_i \}_{i=1}^{N_l}$, and the lane connectivity $\mathbf{A} \in \{0,1\}^{N_l\times N_l \times 4}$. Separate 2-layer MLPs $f_l$ and $f_o$ process the lane and object vectors, respectively, where the lane type and object type are concatenated prior to the MLP
\begin{align*}
    \mathbf{l}^0_i &= f_l([\mathbf{l}_i, c^l_i]), \\ 
    \mathbf{o}^0_i &= f_o([\mathbf{o}_i, c^o_i]),
\end{align*}
where $c^l_i$ is the lane type of lane $i$ and $c^o_i$ is the object type of object $i$. A 2-layer MLP $f_a$ additionally processes the lane connectivity type $a_{ij} := \mathbf{A}[i,j]$ between all (directed) edges $(i,j)$ connecting lane $i$ to lane $j$
\begin{align*}
    \mathbf{a}^0_{ij} &= f_a(a_{ij}).
\end{align*}
The lane, object, and lane connectivity embeddings are then processed by a sequence of $N_E$ factorized attention blocks (FABs), where each FAB consists of sequential lane-to-lane (L2L), lane-to-object (L2O), and object-to-object (O2O) attention layers. First, we define $\mathbb{L}^k := \{ \mathbf{l}^k_i \}_{i=1}^{N_l}$, $\mathbb{O}^k := \{ \mathbf{o}^k_i \}_{i=1}^{N_o}$, and $\mathbb{A}^k := \{ \mathbf{a}^k_{ij} \}_{(i,j) \in [N_l] \times [N_l]}$ as the set of embeddings for the lanes, objects, and lane connectivities following the output of the $k$'th FAB. The $k$'th FAB updates the $k$'th layer lane, object, and lane connectivity embeddings as
\begin{align*}
    (\mathbb{L}^{k+1}, \mathbb{O}^{k+1}, \mathbb{A}^{k+1}) = \text{FAB}_{k}(\mathbb{L}^{k}, \mathbb{O}^{k}, \mathbb{A}^{k}),
\end{align*}
where $\text{FAB}_k$ is decomposed as follows
\begin{align*}
    \mathbb{L}^{k+1} &= \text{L2L}_k(\mathbb{L}^{k}, \mathbb{A}^{k}), \\
    \mathbb{O}^{k+1} &= \text{L2O}_k(\text{proj}^k(\mathbb{L}^{k+1}), \mathbb{O}^{k}), \\
    \mathbb{O}^{k+1} &= \text{O2O}_k(\mathbb{O}^{k+1}), \\
    \mathbb{A}^{k+1} &= \text{EdgeUpdate}_k(\mathbb{L}^{k+1}, \mathbb{A}^{k})
\end{align*}
where L2L is a Transformer encoder block \citep{vaswani2017attention} where the multi-head self-attention operation fuses the edge features $\mathbb{A}^k$ into the keys and values, as in \citep{zhou2023qcnet}. L2O is a Transformer decoder block where the lane embeddings are projected to the object hidden dimension with a linear layer $\text{proj}_k$, and O2O is a Transformer encoder block. As we are processing sets of elements and wish to preserve permutation equivariance, each of the L2L, L2O, and O2O blocks do not have positional encodings. $\text{EdgeUpdate}_k$ updates each lane connectivity embedding $\mathbf{a}^k_{ij}$ as
\begin{align*}
    \mathbf{a}^{k+1}_{ij} = f_k^2([\mathbf{a}^{k}_{ij}, f^1_k([\mathbf{l}^{k+1}_{i}, \mathbf{l}^{k+1}_{j}])]),
\end{align*}
where $[\cdot, \cdot]$ denotes concatenation along the feature dimension, and $f_k^1, f_k^2$ are each 2-layer MLPs. For the partitioned scenes, to predict the number of lanes in $\mathcal{I}_{F_P}$, we additionally define a learnable query vector $\mathbf{q}$ and each $\text{FAB}_k$ in the encoder is augmented with a lane-to-query (L2Q) attention layer following the L2L layer
\begin{align*}
    \mathbb{Q}^{k+1} = \text{L2Q}_k(\mathbb{L}^{k+1}_{F_N}, \mathbb{Q}^k),
\end{align*}
where $\mathbb{Q}^0 = \{ \mathbf{q} \}$ and $\mathbb{L}^{k+1}_{F_N}$ denotes the embeddings of the lanes in $\mathcal{I}_{F_N}$ of a partitioned scene.

After $N_E$ FABs, the lane and object embeddings are projected to mean and variance parameters with latent dimension $K_l$ and $K_o$, respectively, as in the standard VAE, and the resulting transformed query vector $\mathbf{q}^{N_E}$ is passed through a 3-layer MLP $f_{\text{num}}$ to predict the number of lanes in $\mathcal{I}_{F_P}$
\begin{align*}
    \boldsymbol{\mu}^i_{\mathcal{L}} &= f^{{\mathcal{L}}}_{\text{mean}}(\mathbf{l}^{N_E}_i), \\
    \boldsymbol{\sigma}^i_{\mathcal{L}} &= f^{{\mathcal{L}}}_{\text{std}}(\mathbf{l}^{N_E}_i), \\
    \boldsymbol{\mu}^i_{\mathcal{O}} &= f^{\mathcal{O}}_{\text{mean}}(\mathbf{o}^{N_E}_i), \\
    \boldsymbol{\sigma}^i_{\mathcal{O}} &= f^{\mathcal{O}}_{\text{std}}(\mathbf{o}^{N_E}_i),
    \\
    \hat{N}^{F_P}_l &= f_{\text{num}}(\mathbf{q}^{N_E})
\end{align*}
where $f^{\mathcal{L}}_{\text{mean}}, f^{\mathcal{L}}_{\text{std}}, f^{\mathcal{O}}_{\text{mean}}, f^{\mathcal{O}}_{\text{std}}$ are linear layers.

\textbf{Decoder} Given sampled latents $\{ \{\mathbf{h}^{\mathcal{O}}_i\}_{i=1}^{N_o},  \{\mathbf{h}^{\mathcal{L}}_i\}_{i=1}^{N_l} \} \sim \mathcal{E}_{\phi}$ from the encoder, separate 2-layer MLPs $f_{h_l}$ and $f_{h_o}$ process the lane and object latents, respectively
\begin{align*}
    \mathbf{l}^{0}_i &= f_{h_l}(\mathbf{h}^{\mathcal{L}}_i), \\
    \mathbf{o}^{0}_i &= f_{h_o}(\mathbf{h}^{\mathcal{O}}_i).
\end{align*}
We additionally derive lane connectivity features $\mathbf{a}^0_{ij}$ by processing the incident lane embeddings $\mathbf{l}^{0}_i, \mathbf{l}^{0}_j$ through a 2-layer MLP $f_{h_a}$ for all $i,j$
\begin{align*}
    \mathbf{a}^0_{ij} &= f_{h_a}([\mathbf{l}^{0}_i, \mathbf{l}^{0}_j])
\end{align*}
Then, as in the encoder, we process the lane, object, and lane connectivity embeddings with a stack of $N_D$ FABs. The resulting embeddings are used to predict the lane positions, lane type, object features, object type, and lane graph connectivity, for all $i,j$
\begin{align*}
    \hat{\mathbf{l}}_i &= f_{\text{lane}}(\mathbf{l}^{N_D}_i), \\
    \hat{c}_i^l &= f_{\text{lane-type}}(\mathbf{l}^{N_D}_i), \\
    \hat{\mathbf{o}}_i &= f_{\text{object}}(\mathbf{o}^{N_D}_i), \\
    \hat{c}_i^o &= f_{\text{object-type}}(\mathbf{o}^{N_D}_i), \\
    \hat{a}_{ij} &= f_{\text{connectivity}}(\mathbf{a}^{N_D}_{ij}),
\end{align*}
where $f_{\text{lane}}, f_{\text{object}}$ are 4-layer MLPs and $f_{\text{lane-type}}, f_{\text{object-type}}, f_{\text{connectivity}}$ are 3-layer MLPs.

The autoencoder loss function is defined by
\begin{align*}
   L_{\text{ae}} &= \underbrace{ \lambda_{\text{lane}} \frac{1}{N_l} \sum_{i=1}^{N_l} \Big( \ell_2(\mathbf{l}_i, \hat{\mathbf{l}}_i) + \text{ce}(\hat{c}_i^l, c_i^l) \Big) + \frac{1}{N_o} \sum_{i=1}^{N_o} \Big( \ell_2(\mathbf{o}_i, \hat{\mathbf{o}}_i) + \text{ce}(\hat{c}_i^o, c_i^o) \Big) +  \lambda_{\text{conn}} \frac{1}{N_l^2} \sum_{i = 1}^{N_l}\sum_{j = 1}^{N_l} \text{ce}(\hat{a}_{ij}, a_{ij})}_{\text{reconstruction loss}} \\
   &- \underbrace{\frac{\beta}{2} \Big( \frac{1}{N_l} \sum_{i=1}^{N_l} \Big[ 1 + \log(\boldsymbol{\sigma}^{\mathcal{L}}_i)^2 - (\boldsymbol{\mu}^{\mathcal{L}}_i)^2 - (\boldsymbol{\sigma}^{\mathcal{L}}_i)^2 \Big] + \frac{1}{N_o} \sum_{i=1}^{N_o} \Big[ 1 + \log(\boldsymbol{\sigma}^{\mathcal{O}}_i)^2 - (\boldsymbol{\mu}^{\mathcal{O}}_i)^2 - (\boldsymbol{\sigma}^{\mathcal{O}}_i)^2 \Big] \Big)}_{\text{KL loss}}
   + \lambda_{\text{num}} \text{ce}(\hat{N}_l^{F_P}, N_l^{F_P}),
\end{align*}
where $\lambda_{\text{lane}}, \lambda_{\text{conn}}, \lambda_{\text{num}}$ are coefficients to scale the respective lanes, $\text{ce}(\cdot, \cdot)$ denotes the cross entropy loss, and $\beta$ scales the KL loss. For non-partitioned scenes, $\lambda_{\text{num}} = 0$.  

\subsubsection{Latent Diffusion Model}

The latent diffusion model models the joint distribution over lane and object latents: $p_{\theta}(\{ \mathbf{h_i^{\mathcal{L}}} \}_{i=1}^{N_l}, \{ \mathbf{h_i^{\mathcal{O}}}_{i=1}^{N_o} \} | N_o,N_l)$. We let $\mathbf{H}_0$ denote the matrix of stacked lane and object latents. Then, the diffusion model models the distribution: $p_{\theta}(\mathbf{H}_0 | N_o,N_l)$.

\textbf{Forward Diffusion Process} We define a forward noising process over $T$ steps for the latents $\mathbf{H}_0$. Following DDPM \citep{ho2020ddpm}, starting from data $\mathbf{H}_0 \sim q(\mathbf{H}_0)$, we can define a chain of noisy latents with a variance schedule $\beta_1, \dots, \beta_T$
\begin{align*}
    q(\mathbf{H}_{1:T} | \mathbf{H}_{0}) &= \prod_{t=1}^T q(\mathbf{H}_t | \mathbf{H}_{t-1}), \\ q(\mathbf{H}_t | \mathbf{H}_{t-1})
    &:= \mathcal{N}(\mathbf{H}_{t}; \sqrt{1 - \beta_t} \mathbf{H}_{t-1}, \beta_t \mathbf{I}), 
\end{align*}
Leveraging the properties of Gaussians, we have that
\begin{align}
    \label{gaussian}
     q(\mathbf{H}_t | \mathbf{H}_0) &= \mathcal{N}(\mathbf{H}_t; \sqrt{\bar{\alpha}_t} \mathbf{H}_0, (1 - \bar{\alpha}_t)\mathbf{I}),
\end{align}
where $\alpha_t := 1 - \beta_t$ and $\bar{\alpha}_t := \prod_{s=1}^t \alpha_s$. 

\textbf{Reverse Diffusion Process} We define a corresponding reverse diffusion process for the latents. Given a sufficiently large $T$, the distribution of latents is approximately distributed as $\mathbf{H}_T \sim \mathcal{N}(\boldsymbol{0}, \mathbf{I})$. If we can sample from $q(\mathbf{H}_{t-1} | \mathbf{H}_t)$, then we can sample from $q(\mathbf{H}_0)$ by first sampling noise $\mathbf{H}_T \sim \mathcal{N}(\boldsymbol{0}, \mathbf{I})$, and iteratively sampling from $\mathbf{H}_{t-1} \sim q(\mathbf{H}_{t-1} | \mathbf{H}_t)$ for $T$ steps. Unfortunately, $q(\mathbf{H}_{t-1} | \mathbf{H}_t)$ is intractable and thus we approximate it with a neural network. Concretely, we define a Markov chain starting from $p(\mathbf{H}_T) = \mathcal{N}(\boldsymbol{0}, \mathbf{I})$
\begin{align*}
    p_{\theta}(\mathbf{H}_{0:T}) &:= p(\mathbf{H}_{T}) \prod_{t=1}^T p_{\theta}(\mathbf{H}_{t-1} | \mathbf{H}_t), \\
    p_{\theta}(\mathbf{H}_{t-1} | \mathbf{H}_t) &= \mathcal{N}(\mathbf{H}_{t-1}; \boldsymbol{\mu}_{\theta}(\mathbf{H}_{t}, t), \Sigma_{t}).
\end{align*}
Following \citep{ho2020ddpm}, we let $\Sigma_t := \tilde{\beta}_t \mathbf{I} := \frac{1 - \bar{\alpha}_{t-1}}{1 - \bar{\alpha}_t} \beta_t$, and we utilize the $\epsilon$-parameterization of $\boldsymbol{\mu}_{\theta}(\mathbf{H}_{t}, t)$:
\begin{align*}
    \boldsymbol{\mu}_{\theta}(\mathbf{H}_{t}, t) := \frac{1}{\sqrt{\alpha_t}} (\mathbf{H}_{t} - \frac{1 - \alpha_t}{\sqrt{1 - \bar{\alpha}_t}} \epsilon_{\theta}(\mathbf{H}_{t}, t)).
\end{align*}
\textbf{Training Objective} To train $\epsilon_{\theta}$, we optimize the variational lower bound to the log-likelihood (ELBO)
\begin{align}
\label{vlb}
    - \mathbb{E}_{q(\mathbf{H}_{0})} \log p_{\theta}(\mathbf{H}_{0}) \leq \mathbb{E}_{q(\mathbf{H}_{0:T})}[\log \frac{q(\mathbf{H}_{1:T}|\mathbf{H}_{0})}{p_{\theta}(\mathbf{H}_{0:T})}].
\end{align}
Up to reweighting, optimizing Equation \ref{vlb} is equivalent to optimizing the simple DDPM objective
\begin{align}
    \label{obj1}
    L_{\text{dm}} &= \mathbb{E}_{t, \mathbf{H}_{t}, \boldsymbol{\epsilon}_{t}}\Big[ \boldsymbol{ || \epsilon}_{t} - \epsilon_{\theta}(\mathbf{H}_{t}, t) ||^2_2\Big] \\
    \label{obj2}
    &= \mathbb{E}_{t, \mathbf{H}_{0}, \boldsymbol{\epsilon}_{t}}\Big[ \boldsymbol{ || \epsilon}_{t} - \epsilon_{\theta}(\sqrt{\bar{\alpha}_t}\mathbf{H}_{0} + \sqrt{1 - \bar{\alpha}_t} \boldsymbol{\epsilon}_t, t) ||^2_2\Big],
\end{align}
where (\ref{obj2}) is derived from an application of (\ref{gaussian}).

\textbf{Architecture} We decompose $\mathbf{H}_t$ into lane latents $\mathbf{H}^{\mathcal{L}}_t$ and object latents $\mathbf{H}^{\mathcal{O}}_t$. The latent diffusion model $\epsilon_{\theta}$ takes as input a set of $N_l$ noised lane latents and $N_o$ noised object latents $(\mathbf{H}^{\mathcal{L}}_t, \mathbf{H}^{\mathcal{O}}_t) := \{ \{ \mathbf{h}^{\mathcal{L}}_{i,t}\}_{i=1}^{N_l},  \{ \mathbf{h}^{\mathcal{O}}_{i,t}\}_{i=1}^{N_o} \}$ and a timestep $t$, and predicts the noise $\boldsymbol{\epsilon}_{t}$ such that $\mathbf{H}_t = \sqrt{\bar{\alpha}_t}\mathbf{H}_{0} + \sqrt{1 - \bar{\alpha}_t} \boldsymbol{\epsilon}_t$. We first embed the noisy lane and object latents with separate 2-layer MLPs for all $i$, where $f_{\text{emb},l}$ embeds the lane latents into hidden dimension $d_l$ and $f_{\text{emb},o}$ embeds the object latents into hidden dimension $d_o < d_l$. We additionally apply an additive sinusoidal positional encoding to each embedded latent vector to mitigate the effects of permutation ambiguity
\begin{align*}
    \mathbf{l}^{0}_i &= f_{\text{emb},l}(\mathbf{h}^{\mathcal{L}}_{i,t}) + \mathbf{p}_i^{\mathcal{L}}, \\
    \mathbf{o}^{0}_i &= f_{\text{emb},o}(\mathbf{h}^{\mathcal{O}}_{i,t}) + \mathbf{p}_i^{\mathcal{O}},
\end{align*}
 where $\mathbf{p}_i^{\mathcal{L}}, \mathbf{p}_i^{\mathcal{O}}$ correspond to the $i$'th lane and object positional encodings, respectively.
 We then process the embedded latents through a sequence of $N_{\text{DM}}$ FABs with AdaLN-Zero conditioning \citep{Peebles2022DiT}. Concretely, each FAB in the latent diffusion model consists of the following operations
 \begin{align*}
     \mathbb{L}^{k+1} &= \text{O2L}_k(\text{proj}^k_1(\mathbb{O}^{k}), \mathbb{L}^{k}, \mathbf{C}), \\
     \mathbb{L}^{k+1} &= \text{L2L}_k(\mathbb{L}^{k+1}, \mathbf{C}), \\
    \mathbb{O}^{k+1} &= \text{L2O}_k(\text{proj}^k_2(\mathbb{L}^{k+1}), \mathbb{O}^{k+1}, \mathbf{C}), \\
    \mathbb{O}^{k+1} &= \text{O2O}_k(\mathbb{O}^{k+1}, \mathbf{C}), 
 \end{align*}
 where $\text{O2L}_k, \text{L2L}_k, \text{L2O}_k, \text{O2O}_k$ are $\text{DiT}$ blocks \citep{Peebles2022DiT} with AdaLN-Zero conditioning on vector $\textbf{C}$, where $\mathbf{C}$ consists of the summation of a sinusoidal positional encoding of the timestep $t$, a learnable embedding encoding the scene type identity (\textit{i.e.,} partitioned or non-partitioned scene), and a learnable embedding encoding the city (\textit{i.e.,} Singapore, Las Vegas, Boston, or Pittsburgh for the nuPlan dataset, and simulation-compatible or simulation-incompatible for Waymo). $\text{proj}^k_1$ projects the embedded object latents into dimension $d_l$, and $\text{proj}^k_2$ projects the embedded lane latents into dimension $d_o$. Following $N_{\text{DM}}$ FABs, the resulting lane and object embeddings are decoded with 4-layer MLPs
 \begin{align*}
     \hat{\boldsymbol{\epsilon}}_{\mathcal{L}, t}^i &= f_{\epsilon}^l(\mathbf{l}^{N_{\text{DM}}}_i), \\
     \hat{\boldsymbol{\epsilon}}_{\mathcal{O}, t}^i &= f_{\epsilon}^o(\mathbf{o}^{N_{\text{DM}}}_i).
 \end{align*}

\subsubsection{Training and Implementation Details}

\begin{figure*}[tb]
    \centering
    \begin{subfigure}[b]{0.3\textwidth}
        \centering
        \includegraphics[width=\textwidth]{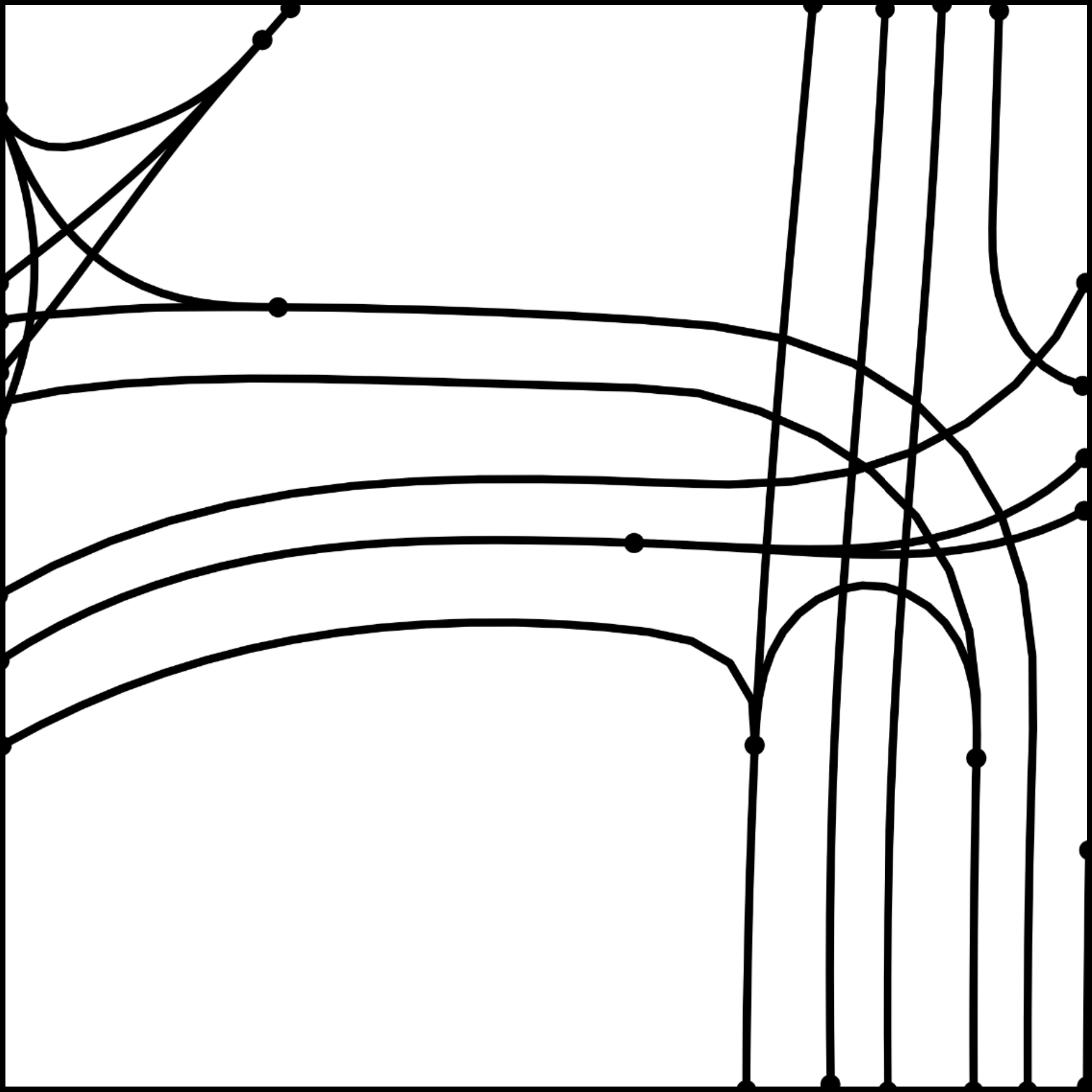}
        \caption{\textbf{Non-partitioned scene}}
        \label{fig:nonpartitioned}
    \end{subfigure}
    \hspace{0.05\textwidth}
    \begin{subfigure}[b]{0.3\textwidth}
        \centering
        \includegraphics[width=\textwidth]{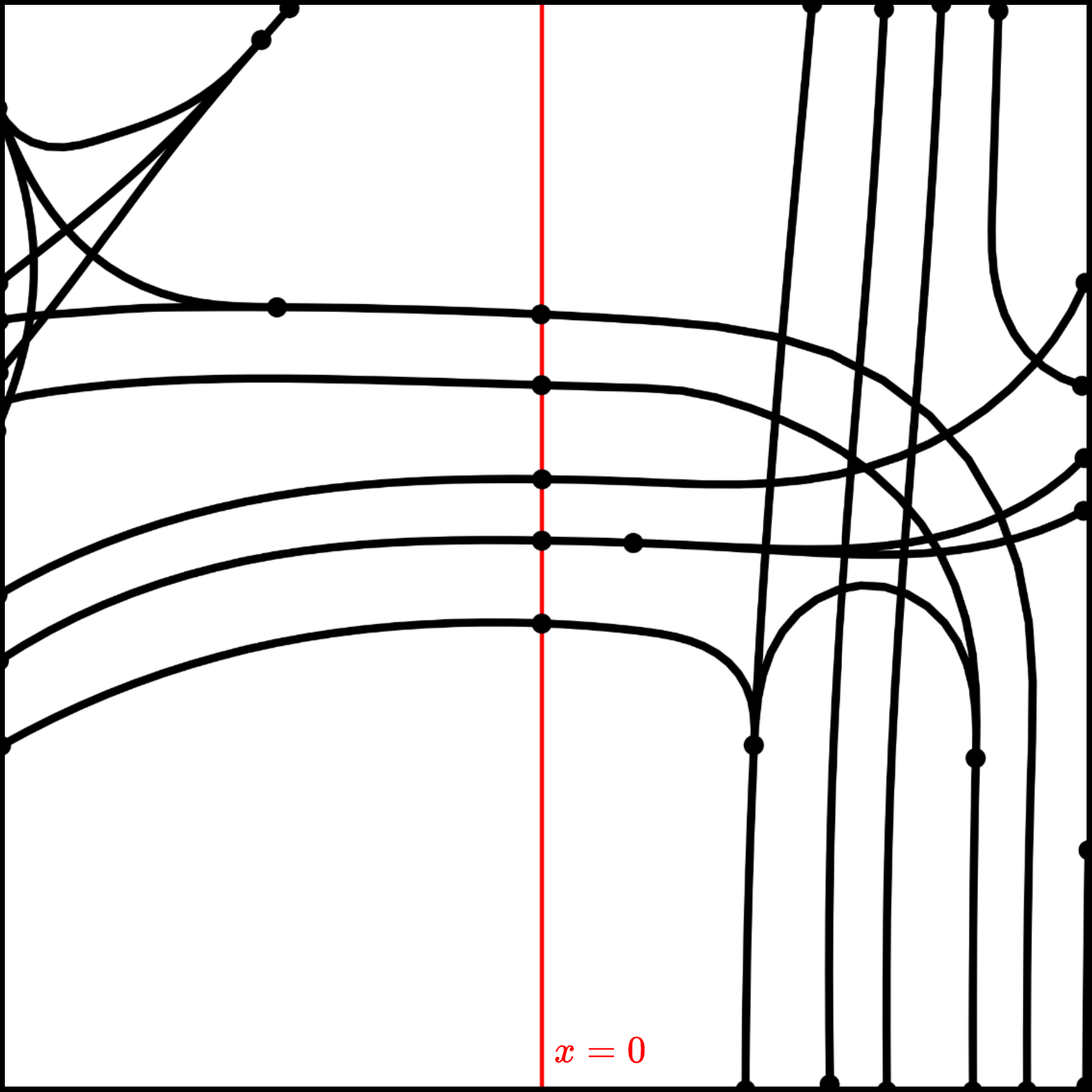}
        \caption{\textbf{Partitioned scene}}
        \label{fig:partitioned} 
    \end{subfigure}
    \caption{\textbf{Partitioned vs. Non-partitioned Scenes.} In Figure \ref{fig:nonpartitioned}, we show a non-partitioned scene where lanes can arbitrarily cross the $x=0$ boundary. In Figure \ref{fig:partitioned}, we split the lanes that cross the $x=0$ boundary (shown in \color{red}red\color{black}).}
    \label{fig:partitionedvsnonpartitioned}\vspace{-15pt}
\end{figure*}

\textbf{Dataset} The nuPlan dataset includes 1,300 hours of logged data across four cities. Following \citet{chitta2024sledge}, we sample 450,000 frames for training and 50,000 for validation, with sampling intervals of 30s, 1s, 2s, and 2s for Las Vegas, Boston, Pittsburgh, and Singapore, respectively. The nuPlan dataset comprises logged scenarios from four cities: Singapore, Las Vegas, Boston, and Pittsburgh. It includes three object types—vehicles, pedestrians, and static objects—and three lane types: centerline, green light, and red light. In the Scenario Dreamer model, traffic light configurations are represented as lane vectors overlapping with the lane centerlines. The Waymo Open Motion Dataset provides $487,002$ training scenarios and $44,097$ validation scenarios, covering 1,750 km of unique roadways. To construct initial scenes for training, we sample each scenario at a random timestep. The Waymo Open Motion dataset contains logged scenarios from six U.S. cities: San Francisco, Phoenix, Mountain View, Los Angeles, Detroit, and Seattle. Each Waymo scene is labeled as either simulation-compatible (\textit{i.e.,} no traffic light signals) or simulation-incompatible following the Nocturne \citep{vinitsky2022nocturne} filtering process. This classification allows us to generate simulation-compatible scenes at inference time via classifier-free guidance while leveraging the simulation-incompatible scenes during training. For the Waymo dataset, the Scenario Dreamer model accounts for three object types—vehicles, pedestrians, and bicycles—and a single lane type representing lane centerlines.

We preprocess both the nuPlan and Waymo datasets into partitioned and non-partitioned scenes. Figure \ref{fig:partitionedvsnonpartitioned} illustrates examples of both formats. For partitioned scenes, we add successor/predecessor connections to lanes split by the \(x=0\) border. Additionally, we follow the lane graph preprocessing approach introduced in SLEDGE \citep{chitta2024sledge}, merging adjacent lanes that permit only a single traversable path. This step effectively removes all key points in the lane graph with degree \(=2\). Each traffic scene is processed within a 64m\(\times\)64m field of view (FOV) centered on the ego agent. For the Waymo dataset, we further exclude off-road vehicles, defined as those located more than 1.5m from a lane centerline.

\textbf{Autoencoder} The architectural configurations for the autoencoder are consistent across datasets, with some dataset-specific adjustments. For the nuPlan dataset, the autoencoder includes an additional 1M parameters to predict the lane type (centerline red or green traffic light), a feature not present in the Waymo dataset as global traffic light configurations are not modelled in the Waymo dataset. The autoencoder is trained for 50 epochs on a single A100 GPU, which takes approximately 36 hours. Dataset-specific limits are set as follows: \(N_o = 30\) and \(N_l = 100\) for Waymo, and \(N_o = 61\) and \(N_l = 100\) for nuPlan. The model uses \(N_E = 2\) encoder and \(N_D = 2\) decoder factorized Transformer blocks. Hidden dimensions are set to 1024 for lanes, 512 for agents, and 64 for lane connectivities. The autoencoder also defines lane and agent latent dimensions as 24 and 8, respectively. Training is conducted using the AdamW optimizer with a learning rate of \(1e^{-4}\), weight decay of \(1e^{-4}\), and a total batch size of 128. The learning rate follows a linear decay schedule with 1000 warmup steps. Hyperparameters include \(\lambda_{\text{lane}} = \lambda_{\text{conn}} = 10\), \(\lambda_{\text{num}} = 0.1\), and \(\beta = 0.01\). The autoencoder features are normalized to $[-1,1]$ by computing the minimum and maximum values of each lane/object attribute in the training dataset. Dropout is not applied during training.

\textbf{Latent Diffusion Model} The latent diffusion model employs \(N_{\text{DM}} = 2\) factorized Transformer blocks, with hidden dimensions of 2048 for lanes and 512 for agents. Following by \citep{rombach2021ldm}, we normalize the latents by calculating the mean and standard deviation from a subset of sampled latents in the training set. During training, we sample these normalized latents to construct each batch, rather than directly regressing on the mean latent as is done in SLEDGE \citep{chitta2024sledge}, which we found introduces a beneficial regularization effect. The base model (B) includes one lane-to-lane (L2L) DiT block within each factorized attention block, while the large model (L) incorporates three L2L blocks, allocating additional capacity to lane generation, which requires greater model capacity than agent generation. Both models use the same autoencoder latents for training. The base and large models are trained with the AdamW optimizer, using a constant learning rate of \(1e^{-4}\), weight decay of \(1e^{-5}\), and a total batch size of 1024. We apply an exponential moving average (EMA) of 0.9999. Lane noise prediction loss is scaled by a factor of 10. The diffusion model employs 100 diffusion steps with a cosine variance schedule (\(\beta_t\)) \citep{nichol2021improvedddpm}. During sampling, low-temperature sampling \citep{ajay2023decisiondiffuser} is applied with \(\alpha = 0.75\) to the lane latents in the reverse diffusion chain. Additionally, noisy lane and object latents \(\mathbf{H}_t\) are clipped to the range \((-5, 5)\) after each denoising step. We dropout the conditioning information during training with probability $0.1$ and apply a classifier guidance scale $s=4.0$ at inference. No dropout is applied.

\begin{figure*}
  \centering
  \includegraphics[width=0.4\textwidth]{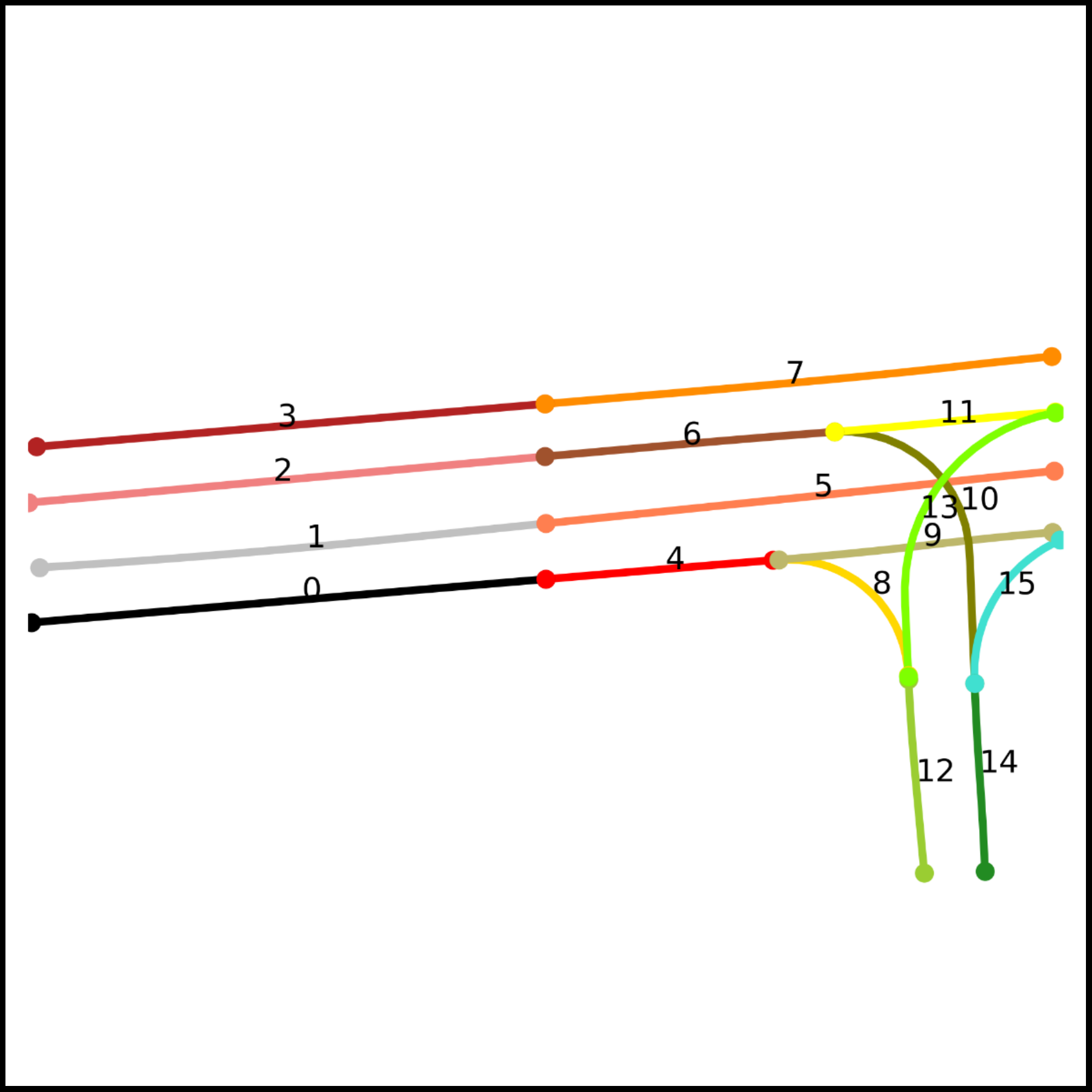}
    \caption{\textbf{Proposed ordering scheme.} We illustrate the proposed ordering scheme for the lanes of a partitioned scene, where agent ordering follows a similar approach. We order first by minimum $x$-value, and if the difference is less than $\epsilon$, we then order by minimum $y$-value, then maximum $x$-value, then maximum $y$-value. Importantly, all lanes before the $x=0$ boundary (in $\mathcal{I}_{F_N}$) are assigned an ordering strictly lower than the lanes after the $x=0$ boundary (in $\mathcal{I}_{F_P}$). }
    \label{fig:ordering}
\end{figure*}

To address the challenges associated with permutation ambiguity during training, we order the lane latents and object latents by minimum $x$-value, and if $x$-values differ by less than $\epsilon=0.5$ meters, they are subsequently ordered by minimum $y$-value, then maximum $x$-value, and finally maximum $y$-value. Figure \ref{fig:ordering} illustrates the proposed ordering for the lanes and agents within a given scene. Ordered as above, a sinusoidal positional encoding is then applied to the embedded lane and object latents to imbue the noisy latents with relative spatial information.

\subsection{CtRL-Sim Behaviour Model}

\subsubsection{Architectural Details}

The Scenario Dreamer behaviour model builds upon the architectural details of CtRL-Sim \citep{rowe2024ctrlsim}, with modifications to support arbitrarily long rollouts without relying on log-replay trajectories. In CtRL-Sim, the goal state for each agent is defined as the final state in the log-replay trajectory and used as conditioning. However, since the Scenario Dreamer initial scene generator provides only the initial agent states, we remove goal conditioning and instead model agent behaviours directly from these initial states. CtRL-Sim employs an encoder-decoder Transformer architecture for multi-agent behaviour simulation. The initial scene generated by Scenario Dreamer is encoded using an encoder with \(E\) Transformer encoder blocks. The decoder then processes tokenized trajectory sequences \(x = \langle \dots, s^1_t, G^{1}_t, a^1_t, \dots, s^N_t, G^{N}_t, a^N_t, \dots \rangle\) with \(D\) Transformer decoder blocks, utilizing a temporally causal mask. Here, \(s_t^i\), \(G_t^i\), and \(a_t^i\) represent the state, return, and action of agent \(i\) at timestep \(t\), with \(N\) denoting the number of agents in the scene.

\begin{figure*}[t]
  \centering
  \includegraphics[width=0.9\textwidth]{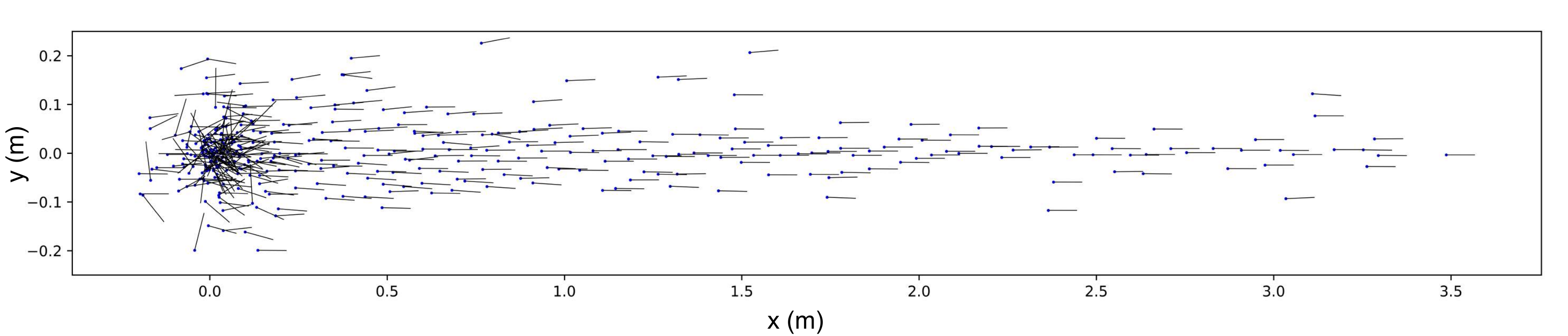}
  \caption{\textbf{Visualization of the $k$-disks tokenization scheme.} Each point denotes the $\Delta x, \Delta y$ offset and $\Delta \theta$ offset denoted by the extending line.}
  \label{fig:kdisks}
\end{figure*}

To accommodate multiple agent types (e.g., vehicles, pedestrians, cyclists), we adopt the $k$-disks tokenization scheme \citep{philion2024trajeglish}, utilizing a discrete vocabulary of size 384, as visualized in Figure \ref{fig:kdisks}. To enable extended rollouts, we also remove absolute timestep conditioning, ensuring that the model can generalize over varying time horizons. Since CtRL-Sim simulates agent behaviours on top of Scenario Dreamer-generated maps, we adapt its map encoder to process Scenario Dreamer maps in a compact representation. In this representation, adjacent lanes are merged when there is only one traversable path, simplifying the lane structure. Specifically, CtRL-Sim operates on a map consisting solely of lane centerlines, with additional map elements such as crosswalks, stop signs, and road edges excluded.

The original CtRL-Sim models the undiscounted return over fixed trajectory lengths \(T\), defined as \(G_t := \sum_{i=t}^T r_i\). However, this approach does not generalize to trajectories longer than \(T\), limiting its applicability to rollouts of arbitrary length. To enable CtRL-Sim to support rollouts longer than those seen during training, we instead model the \textit{discounted return} over a fixed horizon of \(H = 2\) seconds, defined as \(G_t := \sum_{i=t}^{t+H} r_i\). This adjustment enhances controllability in practice, as the discounted return focuses on a shorter, more actionable horizon of 2 seconds, and this horizon better reflects the decision-making timescales relevant to driving.

\textbf{Reward function} As we wish to explicitly aim to produce agent behaviours that challenge the planner, we define the reward function as follows
\begin{align*}
    r_t(s_i^t, s_{\text{ego}}^t) &= -10 \times \mathbbm{1}_{\text{veh-ego coll}}(s_i^t) + \frac{|| s_i^t - s_{\text{ego}}^t ||_2, 10)}{10},
\end{align*}
where $s_{\text{ego}}^t$ is the state of the ego vehicle, or planner, at timestep $t$, and $\mathbbm{1}_{\text{veh-ego coll}}(s_i^t)$ is an indicator function for if vehicle $i$ at timestep $t$ is colliding with the ego.

\textbf{Controllability} The primary advantage of CtRL-Sim over a purely imitation-learning based approach to behaviour simulation is that we can flexibly control agent behaviours at test-time via exponential tilting. Concretely, rather than sampling returns from the learned return distribution $p_{\theta}(G_t|s_t)$, we instead can sample the returns from the exponentially tilted distribution $\return'_t \sim p_\theta (\return_t | \jointstate_t) 
\exp(\kappa \return_t)$, where $G'_t$ is the tilted return-to-go and where $\kappa$ represents the inverse temperature. Higher values of $\kappa$ increase the return density around the best outcomes or higher returns, while negative values of $\kappa$ increase the return density around less favourable outcomes or lower returns.

\subsubsection{Training and Inference Details}

\textbf{Training} We train the CtRL-Sim behaviour model on the full Waymo Open Motion dataset, consisting of $487,002$ training scenarios and $44,097$ validation scenarios; however, we omit a subset of simulation-compatible validation scenarios for testing. The CtRL-Sim behaviour simulation model trains on sequences of length \(H \times N \times 3\), where \(H = 32\) represents the horizon length, and \(N = 24\) is the number of agents. Return-to-go components (\(G_t^{i}\)) are discretized into 350 bins. Following prior work \citep{ngiam2022scenetransformer, philion2024trajeglish, rowe2023fjmp}, the agents and map context are encoded in a global frame by centering and rotating the scene to the ego agent. The agent states are formatted in the same way as object states in Scenario Dreamer; however, we additionally include an binary indicator for the ego vehicle in the scene. The map context is represented by road segments, \(\mathcal{M} := \{\mathbf{l}_i\}_{i=1}^{N_l}\), where each segment \(\mathbf{l}_i := (p_l^1, \dots, p_l^P)\) comprises a sequence of 50 points. We note that as Scenario Dreamer outputs sequences of 20 points, we upsample the Scenario dreamer lane segments to 50 points so that it can be processed by CtRL-Sim. Up to 100 lane segments within an $125$m$\times125$m FOV centered on the ego at a random timestep $t$ in the context are selected as map context, while the social context includes up to 24 agents within a $80$m$\times80$m FOV centered on the ego at timestep $t$. The model employs a hidden dimension \(d = 256\), \(E = 2\) Transformer encoder layers, and \(D = 4\) Transformer decoder layers. We process the trajectories of all dynamic agents, including vehicles, pedestrians, and cyclists. Embeddings for missing states, actions, and returns tokens are set to zero. Additionally, return tokens where the computed return is truncated to horizon $H' < H$ are set to zero. The training process uses the AdamW optimizer with a total batch size of 64, starting with a learning rate of \(5 \times 10^{-4}\), which linearly decays over 500,000 steps. The CtRL-Sim architecture comprises 7.6M parameters and trains in 48 hours on 4 NVIDIA A100 GPUs.

\textbf{Inference} Although CtRL-Sim is trained with a maximum of \(N = 24\) agents, it can handle larger scenes by processing agents in subsets of 24 at each timestep. This process begins by normalizing the scene around the ego agent, and identifying 23 additional agents to form a subset. The first subset includes the 23 agents closest to the ego. The next subset includes the ego and the next 23 agents closest to the ego, and this process repeats iteratively until all agents have been processed. Importantly, we always include the ego vehicle in each 24-agent subset as the return is defined relative to the ego vehicle's position. During inference, the context length is set to \(H = 32\) timesteps, consistent with the training configuration. At each timestep, the 32 most recent timesteps are used as context, with the scene normalized to the ego agent's position and orientation at the latest timestep in the context. Unlike the original CtRL-Sim, which employs a physics-enhanced dynamics model for vehicle behaviour, we utilize a simpler delta-based forward model. This approach applies the predicted \(\Delta x, \Delta y, \Delta \theta\) offsets directly to the agent’s state at each timestep, as derived from the predicted action in the \(k\)-disks vocabulary. While this forward model does not guarantee physical realism, unlike the physics model it is applicable to multiple agent types.

%% file: supplemental/evaluation.tex
\subsection{Metrics}
\subsubsection{Urban Planning Metrics}

The Urban Planning metrics measure the distributional realism of the generated lane graph connectivity by computing $1$-dimensional Frechet distances on node features for nodes with degree $\neq 2$, referred to as \textit{key points} \citep{chu2019turtle, mi2021hdmapgen, chitta2024sledge}. The key points are visualized as black dots in Figure \ref{fig:nonpartitioned}. The following features are computed over $50,000$ ground-truth test lane graphs and generated lane graphs, following the definitions outlined in \citep{chitta2024sledge}:
\begin{itemize}
    \item \textbf{Connectivity}: computes the degrees of all key points across the lane graph. The length of the $1$-d feature list is length $50,000 \times \text{avg-num-keypoints-per-lane-graph}$.
    \item \textbf{Density}: computes the number of key points in each lane graph for a total feature list length of $50,000$.
    \item \textbf{Reach} computes, for each keypoint, the number of paths to other keypoints, for a total feature list length of $50,000 \times \text{avg-num-keypoints-per-lane-graph}$.
    \item \textbf{Convenience} computes the Dijsktra path lengths for all valid paths between key points for a total feature list length of $50,000 \times \text{avg-num-paths-per-lane-graph}$.
\end{itemize}
Following SLEDGE \citep{chitta2024sledge}, we compute the Frechet Distance and \textit{not} the squared Frechet distance. Furthermore, we multiply the Connectivity, Density, Reach, and Convenience Frechet distances by $10$, $1$, $1$, $10$, respectively, for readability. 

\subsubsection{Agent JSD Metrics}

Following SceneControl \citep{lu2024scenecontrol}, we compute the Jensen-Shannon Divergence (JSD) metrics, which evaluate the distributional realism of the initial vehicle bounding box configurations by comparing $50,000$ real and generated scenes. The Jensen Shannon Divergence between two normalized histograms $p$ and $q$ is computed as
\begin{align*}
    \frac{D_{\text{KL}}(p || m) + D_{\text{KL}}(q || m)}{2},
\end{align*}
where $m$ is the pointwise mean of $p$ and $q$ and $D_{\text{KL}}$ is the KL-divergence. We compute the following features to compute JSDs:
\begin{itemize}
    \item \textbf{Nearest Distance} computes the nearest distance between each vehicle and its neighbours. We clip values between $(0,50)$m with a bin size of $1$m. We scale this JSD metric by 10 for readability.
    \item \textbf{Lateral Deviation} computes the lateral deviation to the closest centerline, computed only over vehicles within $1.5$m of a lane centerline. We clip values between $(0,1.5)$m with a bin size of $0.1$m. We scale this JSD metric by 10.
    \item \textbf{Angular Deviation} computes the angular deviation from the closest centerline, computed only over vehicles within $1.5$m of a lane centerline. We clip values between $(-200,200)$ degrees with a bin size of $5$ degrees. We scale this JSD metric by 100.
    \item \textbf{Length} computes the lengths of all vehicles. We clip values between $(0,25)$m with a bin size of $0.1$m. We scale this JSD metric by 100.
    \item \textbf{Width} computes the widths of all vehicles. We clip values between $(0,5)$m with a bin size of $0.1$m. We scale this JSD metric by 100.
    \item \textbf{Speed} computes the speed of all vehicles. We clip values between $(0,50)$m with a bin size of $1$m. We scale this JSD metric by 100.
\end{itemize}
The total feature list length for computing each above JSD metric is $50,000\times \text{avg-num-vehicles-per-scene}$.

\subsubsection{Behaviour Simulation JSD Metrics}

Following CtRL-Sim \citep{rowe2024ctrlsim}, we compute JSD metrics for the following features:
\begin{itemize}
    \item \textbf{Linear speed} computes the speed of all agents at each timestep along the trajectory rollout. We use 200 uniformly spaced bins between 0 and 30 $m/s$.
    \item \textbf{Angular speed} computes the angular speed (change in heading over time) of all agents at each timestep. We use 200 uniformly spaced bins between -50 and 50 degrees.
    \item \textbf{Acceleration} computes the acceleration of all agents at each timestep. We use 200 uniformly spaced bins between -10 and 10.
    \item \textbf{Nearest distance} computes the nearest distance between each vehicle and its neighbours at each timestep. We use 200 uniformly spaced bins between 0 and 40.
\end{itemize}